\documentclass{article}

\usepackage{arxiv}

\usepackage[utf8]{inputenc} 
\usepackage[T1]{fontenc}    
\usepackage{hyperref}       
\usepackage{url}            
\usepackage{booktabs}       
\usepackage{amsfonts}       
\usepackage{nicefrac}       
\usepackage{microtype}      
\usepackage{lipsum}         
\usepackage{graphicx}
\usepackage{natbib}
\usepackage{doi}

\usepackage{amssymb}
\usepackage{amsmath}
\usepackage{multirow}
\usepackage{xcolor}
\usepackage{float}
\usepackage{placeins}
\usepackage{xspace}
\usepackage{cleveref}       

\newcommand{\giovani}[1]{{\color{black} {#1}}}

\newcommand{\myparagraph}[1]{\smallskip\noindent\textbf{#1}\xspace}

\title{Best Practices for Responsible Machine Learning in Credit Scoring}

\newif\ifuniqueAffiliation
\uniqueAffiliationtrue

\ifuniqueAffiliation 
\author{Giovani Valdrighi \\
Institute of Computing\\
University of Campinas\\
Campinas, Brazil \\
\texttt{giovani.valdrighi@ic.unicamp.br} \\
\And
Athyrson M. Ribeiro \\
Institute of Computing\\
University of Campinas\\
Campinas, Brazil \\
\texttt{a203963@dac.unicamp.br} \\
\And
Jansen S. B. Pereira \\
Institute of Computing\\
University of Campinas\\
Campinas, Brazil \\
\texttt{j252477@dac.unicamp.br} \\
\And
Vitoria Guardieiro\thanks{Research developed in a period at University of Campinas.} \\
Department of Computer and Information Science\\
University of Pennsylvania\\
Philadelphia, PA 19104 \\
\texttt{vitoriag@seas.upenn.edu} \\
\And
Arthur Hendricks \\
Institute of Computing\\
University of Campinas\\
Campinas, Brazil \\
\texttt{a217048@dac.unicamp.br} \\
\And
Décio Miranda Filho \\
Institute of Computing\\
University of Campinas\\
Campinas, Brazil \\
\texttt{d236087@dac.unicamp.br} \\
\And
Juan David Nieto Garcia \\
Institute of Computing\\
University of Campinas\\
Campinas, Brazil \\
\texttt{j252842@dac.unicamp.br} \\
\And
Felipe F. Bocca \\
Institute of Computing\\
University of Campinas\\
Campinas, Brazil \\
\texttt{f060642@dac.unicamp.br} \\
\And
Thalita B. Veronese\footnotemark[1] \\
Federal Institute of São Paulo\\
Campinas, Brazil \\
\texttt{thalitabv@ifsp.edu.br} \\
\And
Lucas Wanner \\
Institute of Computing\\
University of Campinas\\
Campinas, Brazil \\
\texttt{wanner@unicamp.br} \\
\And
Marcos Medeiros Raimundo \\
Institute of Computing\\
University of Campinas\\
Campinas, Brazil \\
\texttt{mraimundo@ic.unicamp.br} \\
}
\else
\usepackage{authblk}

\author[1]{{Giovani Valdrighi\thanks{\texttt{giovani.valdrighi@ic.unicamp.br}}}}
\author[1]{{Athyrson M. Ribeiro\thanks{a203963@dac.unicamp.br}}}
\author[1]{{Jansen S. B. Pereira\thanks{j252477@dac.unicamp.br}}}
\author[3]{{Vitoria Guardieiro\thanks{vitoriag@seas.upenn.edu}}}
\author[1]{{Arthur Hendricks\thanks{a217048@dac.unicamp.br}}}
\author[1]{{Décio Miranda Filho\thanks{d236087@dac.unicamp.br}}}
\author[1]{{Juan David Nieto Garcia\thanks{j252842@dac.unicamp.br}}}
\author[1]{{Felipe F. Bocca\thanks{f060642@dac.unicamp.br}}}
\author[2]{{Thalita B. Veronese\thanks{thalitabv@ifsp.edu.br}}}
\author[1]{{Lucas Wanner\thanks{wanner@unicamp.br}}}
\author[1]{{Marcos Medeiros Raimundo\thanks{mraimundo@ic.unicamp.br}}}
\affil[1]{
    Institute of Computing, University of Campinas, Campinas, Brazil
}
\affil[2]{
    Federal Institute of São Paulo, Campinas, Brazil
}
\affil[3]{
    Department of Computer and Information Science, University of Pennsylvania, Philadelphia, PA 19104
}

\fi

\renewcommand{\shorttitle}{\textit{arXiv} Template}

\hypersetup{
pdftitle={A template for the arxiv style},
pdfsubject={q-bio.NC, q-bio.QM},
pdfauthor={David S.~Hippocampus, Elias D.~Striatum},
pdfkeywords={First keyword, Second keyword, More},
}

\begin{document}
\maketitle

\begin{abstract}
The widespread use of machine learning in credit scoring has brought significant advancements in risk assessment and decision-making. However, it has also raised concerns about potential biases, discrimination, and lack of transparency in these automated systems. This tutorial paper performed a non-systematic literature review to guide best practices for developing responsible machine learning models in credit scoring, focusing on fairness, reject inference, and explainability. We discuss definitions, metrics, and techniques for mitigating biases and ensuring equitable outcomes across different groups. Additionally, we address the issue of limited data representativeness by exploring reject inference methods that incorporate information from rejected loan applications. Finally, we emphasize the importance of transparency and explainability in credit models, discussing techniques that provide insights into the decision-making process and enable individuals to understand and potentially improve their creditworthiness. By adopting these best practices, financial institutions can harness the power of machine learning while upholding ethical and responsible lending practices.
\end{abstract}



\newcommand{\figGlobalExp}{
\begin{figure}
    \centering
    \includegraphics[width = \linewidth]{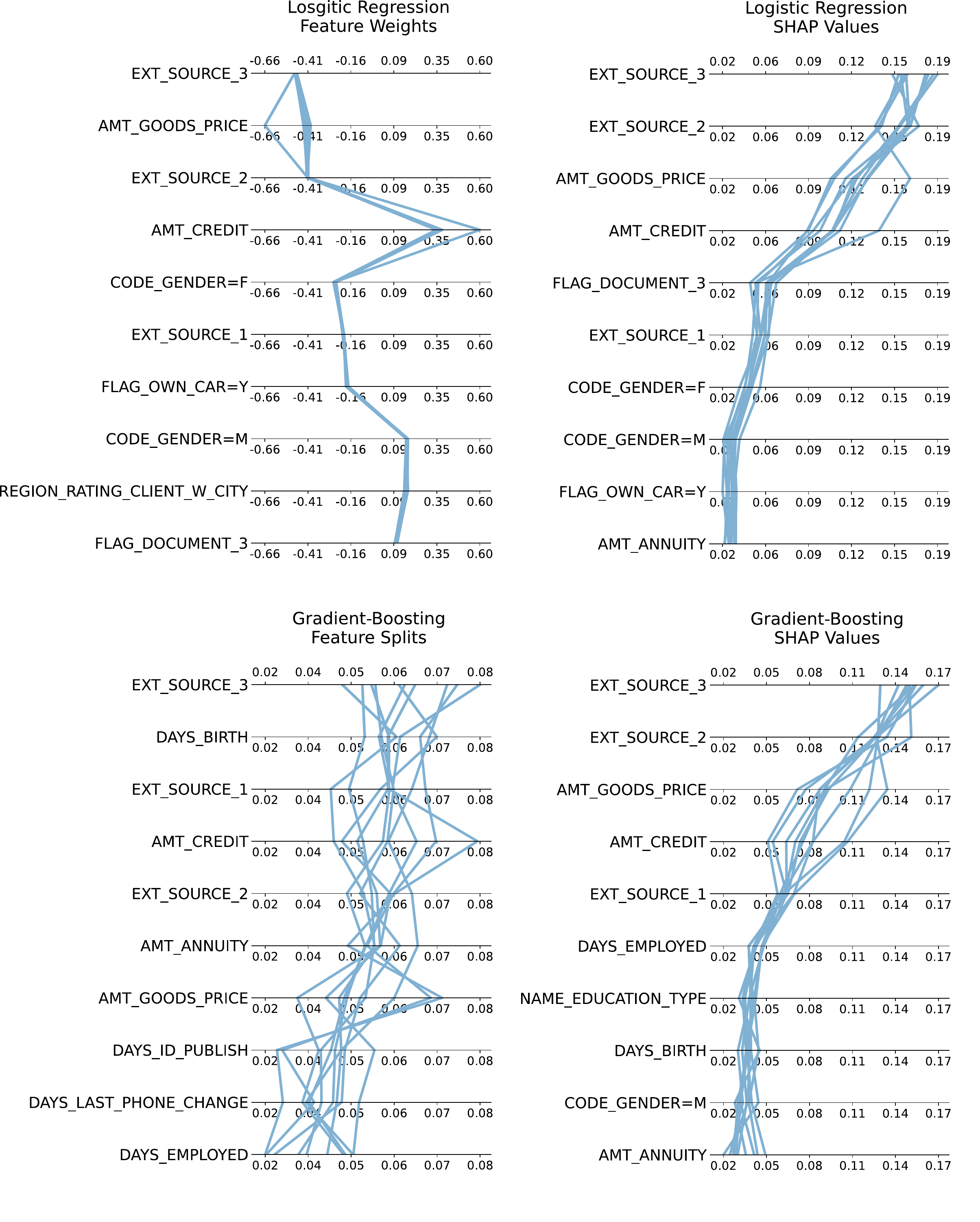}
    \caption{Global explanations calculated for Logistic model and Gradient Boosting at multiple folds using an explainability technique that has access to the model and SHAP values. Displayed features are the 10 with the highest median importance among folds. Credit scores from other institutions (\texttt{EXT\_SOURCE}) were the most important features in almost all models.}
    \label{fig:global_importances}
\end{figure}
}

\newcommand{\figPDPICE}{
\begin{figure}[t]
    \centering
    \includegraphics[width = 0.8\linewidth]{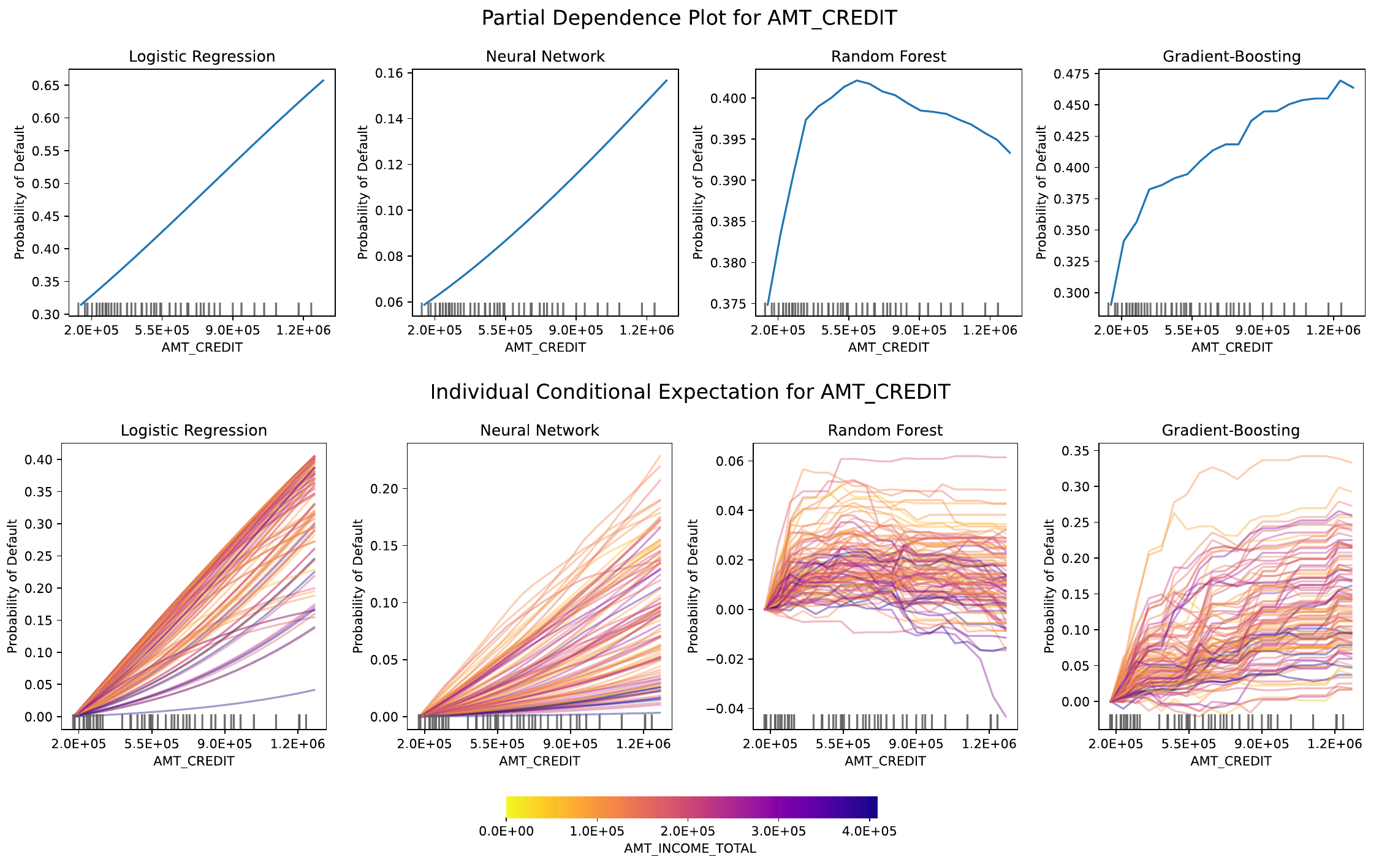}
    \caption{Partial Dependence Plot and Individual Conditional Expectation Plot for four different models with the feature \texttt{AMT\_CREDIT}, that is the total value of requested credit. In the ICE plot, lines are colored based in the feature \texttt{AMT\_INCOME\_TOTAL}, i.e., the client's total income.}
    \label{fig:pdp_ice}
\end{figure}
}

\newcommand{\figSHAPLIME}{
\begin{figure}
    \centering
    \includegraphics[width = 0.8\linewidth]{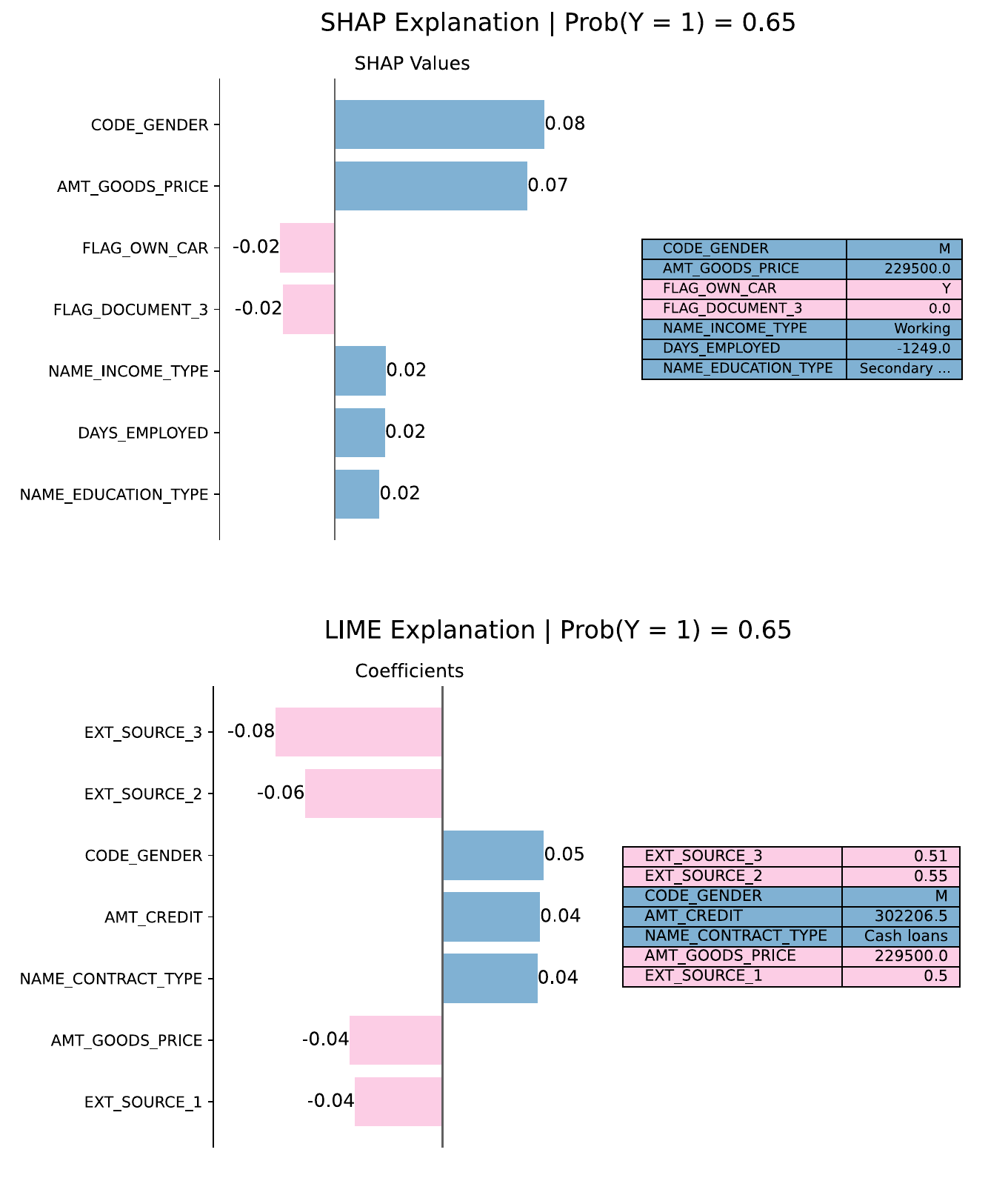}
    \caption{SHAP and LIME local explanations for an individual that was classified with the default outcome. SHAP and LIME explanations can disagree on some occasions.}
    \label{fig:shap_lime}
\end{figure}
}

\section{Introduction}
\label{sec:introduction}

For individuals and families, access to affordable credit is essential as protection against financial volatility, financing and education, pursuing business opportunities, and building equity. From the lender’s perspective, there is a delicate balance between improving access to credit and higher costs due to defaults on payments. Creating responsible credit concession models requires maintaining this balance~\citep{kozodoi2022fairness} while ensuring fair outcomes across different groups of individuals, improving access, and helping applicants understand factors that influence rejection so that they can take action to improve their credit potential.

Credit concession models are created using a variety of data, such as employment history (for example, occupation and income), demographic data (such as age, marital status, and education), and financial data (for example, checking account balance, credit card usage, and bill payment history). Given these features, models such as logistic regression, gradient boosting, and decision trees can be trained to predict whether a new customer will default on a loan over a period of time~\citep{louzada2016classification}. This probability of default can then be used alongside other factors, such as financial exposure, to approve or deny the loan.

Algorithmic and automated decision-making strategies for credit concessions emerged in part due to concerns about the unreliability and discrimination of human decision-makers~\citep {elliehausen1989theory, hand2001modelling}. Algorithmic models can analyze large datasets, identifying hidden patterns from human analysts. However, recent studies demonstrated that these models can perpetuate biases learned from humans or present in the collected data. One common problem identified is the discrimination present in the model outputs based on sensitive attributes of the individual, such as race, religion, or gender. For example, a model might assign a higher credit score if the individual belongs to a race. This problem motivated studies in the area of Fairness, which showed that simply removing sensitive attributes from training data (''fairness through unawareness'') is insufficient to remove biases, as many otherwise non-sensitive features can be correlated and predictive of the sensitive attribute~\citep{barocas2023fairness}. Furthermore, even if there are no proxy features, for some applications this blind approach could miscalibrate predictions and lead to erroneous and potentially harmful decisions \citep{chohlas2023designing}. Methods to improve fairness have been proposed for different steps of the decision-making process, including pre-processing data, in-processing within models, and post-processing outputs.

In addition to biases related to sensitive attributes, credit models suffer from a fundamental exclusionary limitation in that models are based on data from the accepted population, whose performance history is known, rather than the total population. Unbanked individuals, who are generally from disadvantaged social groups, typically lack historical or even recent data on financial status and may experience significant differences in outcome when applying for credit \citep{fuster2022predictably, pager2008sociology}. This means that the models do not accurately represent unbanked and underbanked individuals, as there is no performance data available for them ~\citep{joanes1993reject}. Reject inference methods attempt to incorporate rejected customers into the modeling process by inferring their performance, thus expanding access to previously excluded individuals~\citep{crook2004does, liao2021data}.

Finally, a responsible credit model must allow denied applicants to understand the factors that led to their rejection. For traditional analytical models, this may be simple, but for machine learning models, as model complexity grows, it becomes increasingly difficult to understand how a model made its decision. Explainability techniques can not only pinpoint reasons for rejection, but also suggest actions that an applicant can take to improve their chances of receiving a loan, such as improving the income-to-debt ratio or renegotiating existing debt.

In this work, we explore best practices for creating responsible machine learning models for credit concession. In Sec. ~\ref{sec:credit_models}, we introduce credit concession models and explore commonly used finance datasets. In Sec. ~\ref{sec:fairness}, we introduce the definitions, criteria, and metrics of fairness. We also present different methods to improve fairness by pre-processing data, adjusting models, and correcting biases through post-processing. Then, in Sec. ~\ref{sec:reject_inference}, we introduce techniques to improve access to credit with reject inference, which incorporates data from rejected loan applications into models. Finally, in Sec. ~\ref{sec:explainability}, we introduce methods and metrics for explainability in credit models. Illustrative examples based on open credit datasets are presented throughout the paper, and all examples are supported by the source code available at \url{https://github.com/hiaac-finance}. We conclude the work by discussing open issues and research opportunities.

\section{Machine Learning for Credit}
\label{sec:credit_models}

A fundamental issue for credit modeling is to predict the probability of default, that is, the likelihood that a borrower will fail to repay their loan over a period of time. Credit concession has long included modeling the probability of default as a function of individual characteristics. For example, as early as 1941, \citet{durand1941risk} presented objective credit formulas using different factors based on demographic and personal information. These factors include age, with incremental points awarded for each year of age beyond 20 and up to 50; sex, with greater points allocated for women; occupation, where specific occupations are associated with good-risk or bad-risk; type and stability of employment, with points awarded for working in certain sectors such as banking and government services; and assets such as bank accounts and real estate. These factors were linearly combined, resulting in a composite rating that was then correlated with performance data (good loans and bad loans) for different segments of the ratings. Therefore, this approach allowed a specific rating to predict the probability of default for an individual, which was then used as a guideline, along with other factors, such as loss in the event of default, to approve or deny the loan. 

Numerous alternative approaches and datasets are used to determine creditworthiness. For example, the proprietary FICO algorithm used in consumer lending in the United States calculates scores based on payment history, credit utilization, length of credit history, type of credit accounts, and number of recent credit inquiries. 
Literature reviews by \citet{hand1997statistical}, \citet{louzada2016classification} and \citet{zhang2024consumer} surveyed the statistical and machine learning techniques used for credit scoring in practical applications and research. Methods based on logistic regression~\citep{li2002direct,lee2005two,yap2011using,pavlidis2012adaptive,dong2010credit} are considered the benchmark method for credit scoring.  Additionally, a significant body of work can be found using a variety of methods such as discriminant analysis~\citep{reichert1983examination}, linear regression~\citep{hand2002superscorecards},  k-nearest neighbors (kNN)~\citep{henley1997construction}, decision trees~\citep{yap2011using,kao2012bayesian}, neural networks~\citep{west2000neural,lee2002credit,marcano2011artificial, chuang2011hybrid, tomczak2015classification}, support vector machines (SVM)~\citep{huang2007credit,chen2009mining,li2006evaluation, goh2019credit},  Bayesian networks~\citep{giudici2001bayesian,antonakis2009assessing,okesola2017improved}, as well as hybrid~\citep{dumitrescu2022machine,kyeong2022two} and ensemble~\citep{koutanaei2015hybrid, alaraj2016new} methods. 

Historically, different algorithms have stood out in benchmarks studies (we focus our reading on AUC score): Neural Networks, Radial Basis Function (RBF) SVM, and Logistic Regression in~\citep{baesens2003benchmarking}; Random Forest, Neural Networks and Logistic Regression in~\citep{lessmann2015benchmarking}; Random Forest, and Logistic Regression in~\citep{alaraj2016classifiers}; Gradient Boosting (XGBoost and GBDT), Random Forest, LDA, and Logistic Regression in~\citep{he2018novel}; Ensemble Methods (Bagging, Boosting, and Random Forest) in~\citep{dastile2020statistical} (but Decision Trees and Logistic Regression are also recommended as benchmarks). Using those works as a basis for our study, we narrow our focus to four fundamental methods: logistic regression, neural networks, random forest, and gradient boosting. In the following sections, we present these techniques with more details and the result of an experimental comparison between them.

\subsection{Fundamental Methods}

In this section, we introduce the fundamental methods of logistic regression, neural networks, random forest, and gradient boosting. These models can be used to predict whether a borrower will default on their loan or not. On the following sections, consider the following formulation of the credit scoring problem: with a dataset of individuals $X \in \mathbb{R}^{n \times d}$ and target labels $Y \in \{0, 1\}^n$, where $y_i = 1$ ($i$-th value of $Y$) if the individual represented in $x_i$ defaulted, find a function $f : \mathbb{R}^d \rightarrow [0, 1]$ that approximates $P(Y \mid X)$, i.e., minimizes a loss function $\mathcal{L}(f(x), P(y\mid x))$.

\textbf{Logistic Regression (LR)} models use a sigmoid curve, the logistic function, to estimate the probability of a binary outcome given a set of predictor variables. In a credit model based on logistic regression, the probability of default is modeled as $P(Y=1 | X) = 1/(1 + \exp{(-z)})$, where $z$ is a linear combination of the predictor variables and their corresponding coefficients: $z_i = \beta_{0} + \beta_1x_{i1} + \beta_2x_{i2} + \dots + \beta_dx_{id}$.

Coefficients in the linear combination are estimated using methods like Maximum Likelihood Estimation (MLE). Categorical variables, such as those representing gender, race, marital status, or education level, must be converted to numeric values before they can be used in a logistic model. This is typically done through one-hot encoding or dummy encoding, where categorical variables are converted into a set of new binary variables that indicate the inclusion or exclusion of a specific category.

The models' output, which ranges from 0 to 1, is interpreted as the probability that the input belongs to the positive class, which in the credit model corresponds to client default. For example, if the output is 0.8, it can be interpreted as a default risk of 80\%. The output can be transformed into a binary label with a specified threshold, which is typically set to $0.5$. If the output is greater than or equal to the threshold, the prediction is classified as 1 (default), and if the output is less than the threshold, the prediction is classified as 0 (non-default). Logistic regression models have the benefit of being interpretable, as one can easily identify and understand the most important features to predict the target, as we discuss in Sec.~\ref{sec:explainability}. This method is widely used as a benchmark in credit scoring and default prediction~\citep{louzada2016classification}. 

\textbf{Artificial Neural Networks (ANN)} are built by node components that are loosely inspired by neurons of the human brain. Nodes are organized in layers, and each node is connected to all nodes of the following layers. Each of these connections is associated with a weight. Each node determines its output based on the weighted sum of its inputs, which is added to a bias and fed into an activation function. A typical activation function is ReLU (rectified linear unit), for which the output signal is zero if the weighted input is less than a certain threshold. Otherwise, the output signal is equal to the input signal. The number of layers and the number of nodes in each layer are characteristics of the ANN architecture and are selected differently according to the application. Early work by \citet{west2000neural} compared different credit scoring models based on neural networks. Other approaches include using Multi-Layer Perceptron networks~\citep{zhao2015investigation}, and hybrid models~\citep{lee2002credit, chuang2011hybrid}.

\textbf{Random Forest (RF)} is an ensemble model that relies on a set of decision trees for classification and regression. Each tree is composed of nodes that represent a decision based on a feature and branches that can end on other nodes or an outcome. The decision criterion is typically a threshold that is applied to a single feature in the input vector. The feature and the threshold are chosen to maximize the information gain at the node. While individual decision trees are expressive and easily interpretable, they often suffer from overfitting and low accuracy~\citep{bramer2007principles}. Random forest uses an ensemble strategy that leverages multiple trees, each trained on a subset of data, which are created with replacements so that each data point can be used to train multiple trees. Moreover, individual trees use a random selection of features to decide on the split of each node. This introduces randomness and helps decorrelate the trees. The final decision is made by taking the plurality vote across the individual classification trees. Random forest is more robust to overfitting, generally performs better than individual trees~\citep{breiman2001random}, and can achieve good performance for default prediction~\citep{zhan2018novel}.

\textbf{Gradient boosting (GBM)} is another ensemble of classifiers, which are typically decision trees. However, unlike in random forest where trees are trained individually from different random samples and features, gradient boosting trains a sequence of trees, where each new ``weak'' tree is designed to correct the error of the current ``strong'' model, which is the ensemble of trees built up to this point. Intuitively, at each step, the error of the strong model is calculated, and a new weak model is trained to predict this error. The new weak model is then integrated with the strong model, and this process is repeated for several iterations. The error can be understood as the difference between the prediction and the label. In general, error is defined as a differentiable loss function of the strong model output, such as cross-entropy or average squared error, and the weak model is trained to predict the gradient of the loss. \citet{liu2021step,liu2022credit} introduced credit scoring methods based on Gradient Boosting. Efficient implementations and variations of the basic algorithm include eXtreme Gradient Boosting (XGBoost)~\citep{chen2016xgboost}, LightGBM~\citep{ke2017lightgbm}, and CatBoost~\citep{prokhorenkova2018catboost}.  

\subsection{Performance Metrics}

In an ideal classifier, all good payers would be classified correctly with $0$ (true negatives), and defaults would be correctly classified as $1$ (true positives). However, this is rarely achievable, and the updated objective is to obtain a model that maximizes a quality metric. A commonly utilized metric is accuracy, which is the proportion of overall correct predictions. However, this metric is not enough to reflect the model's performance when there is an imbalance in the number of training examples between the classes, as is often the case in credit scoring systems~\citep{kang2021graph}. A model that always predicts the major class will have high accuracy and will not be useful. For that purpose, balanced accuracy is an adaptation of accuracy in which the correct predictions are weighted by the distribution of labels.


Another important consideration is that accuracy (and balanced accuracy) is only calculated after a classification threshold is selected. This limitation motivates the definition of the Area Under the Curve (AUC) metric:

\begin{equation}
  \textrm{AUC} = P[p(y=1|x_i)>p(y=1|x_j)|y_i=1, y_j=0]
\end{equation}

This metric determines the probability that the score of a sample $i$ belonging to class 1 ($y_i=1$) has its score $p(y|X_i)$ higher than the score $p(y|X_j)$ of a sample $j$ belonging to class 0 ($y_j = 0$).  The higher the AUC value, the better the classifier~\citep{dastile2021making}. Beyond being more robust to data imbalances, the AUC is better designed to deal with different operation points (thresholds for decision). This ranking is fundamental in credit scoring because it serves not just to categorize borrowers but also to rank them, enabling finer interest rate distinctions. This ranking extends beyond product-specific definitions of "good" and "bad" payers, highlighting the importance of AUC (Area Under the Receiver Operating Characteristic Curve) as a key metric in evaluating the efficacy of credit scoring models for accurate risk stratification and pricing.

\begin{table}[t]
\centering
\begin{tabular}{@{}lllll@{} }
\toprule
Dataset  & \# samples & \# features & Target \\ \midrule
Home Credit &  300,000    & 121     & non-default/default       \\
Taiwan      &  30,000     & 24      & credible/non-credible client       \\
German      &  1,000      & 20      & good/bad borrower       \\ \bottomrule
\end{tabular}
\caption{Overview of credit scoring datasets used in experiments.}
\label{tab:datasets}
\end{table}

\subsection{Datasets}
\label{sec:datasets}

Public, comprehensive, and high-quality financial databases are scarce, mainly due to regulatory and privacy requirements \citep{assefa2021generating, bhatore2020machine}. In this work, we use three commonly used publicly available datasets, described below. Datasets are also summarized in Tab.~\ref{tab:datasets}. \giovani{All datasets included information about the individual's gender, which was excluded from the model training due to legal restrictions on its use as a criterion for credit scoring.}

\textbf{Home Credit} dataset~\citep{homecredit} comprises 300,000 training instances representing borrowers' historical data, together with a binary class attribute rating the applicant as either a non-default or a default (after 2 years), split 93.3\% to 6.7\%, respectively. Each instance contains a set of 121 attributes, including 106 numerical and 15 categorical, discrete-valued attributes, the latter having 2 to 58 possible values. 

\textbf{Taiwan} dataset~\citep{taiwan} is composed of 30,000 instances representing Taiwanese bank customers' payment data, together with a binary class attribute rating the applicant as either a credible or a non-credible client, split 78\% to 22\%, respectively. Each instance contains a set of 24 attributes, including 16 numerical and 8 categorical, discrete-valued attributes, the latter having 2 to 10 possible values. 

\textbf{German Credit} Data~\citep{german} is composed of 1,000 instances representing people who once took credit from a bank in Germany, together with a binary class attribute rating the applicant as either a good or a bad borrower, split 70\% to 30\%, respectively. Each instance is represented by a set of 29 attributes, including 7 numerical and 22 categorical, discrete-valued attributes, the latter having 4 to 10 possible values. 

As the datasets usually need common transformations, such as the removal of missing values, encoding of string variables, and others, we developed a pipeline~\footnote{Available at: \url{https://github.com/hiaac-finance/credit_pipeline}} that contains the overall preprocessing steps and the classifier. This pipeline is important to avoid test data contamination, i.e., information that should be only available in the training subset of data but is available in the test one. This pipeline includes operations such as encoding string features, one-hot encoding, and standard scaler.

\subsection{Credit Scoring Experiments}

An experiment was performed to compare the performance of the different fundamental models in the credit scoring datasets presented. The algorithms of the discussed models are implemented in open-source Python packages and can be easily utilized. In particular, logistic regression and random forest were used from the Scikit-learn~\footnote{https://scikit-learn.org/} package, the neural network is based on Keras~\footnote{https://keras.io/}, and the LightGBM~\footnote{https://lightgbm.readthedocs.io/} implementation of gradient boosting was used. A data preprocessing step was necessary for all datasets to deal with missing values, perform one hot encoding, and rescaling of features. The details of this step are available in the supplemental code.

An important detail in comparing the models is the use of hyperparameter tuning. To achieve that, the data was separated into training, validation, and testing. Optuna~\citep{akiba2019optuna} was utilized to obtain hyperparameter values that achieved the highest AUC scores in the validation set. The hyperparameters optimized for each algorithm were selected based on the literature, and a similar number of hyperparameters was considered for all techniques to achieve comparability of the results.
$100$ trials of hyperparameter optimization were executed for each model (and each dataset).
We fixed the test set to obtain a statistic and utilized 10 random separations of the remaining data in training and validation.
Training data included 80\% samples and validation and test used 10\% each.

Tab.~\ref{tab:credit_models_results} displays the mean AUC scores of each algorithm and dataset.
It is possible to see that the overall performance was between 0.7 and 0.8, without great differences between the datasets.
\giovani{While logistic regression obtained the best result on the smaller dataset, gradient boosting, which is a more complex model, obtained the best results on the larger datasets.
While random forest was surpassed by a small margin on Taiwan and Home Credit datasets by gradient-boosting, its performance was better than gradient-boosting on the German dataset by a significant margin.}
In conclusion, this result illustrates that there is competitiveness among methods, with no one being able to surpass all of the others in any situation.


\begin{table}[t]
\centering
\begin{tabular}{@{}llll@{}}
\toprule
\textbf{Model}      & \textbf{German} & \textbf{Taiwan} & \textbf{Home Credit} \\ \midrule
Logistic Regression & \textbf{0.761} ($\pm$ 0.008) & 0.770 ($\pm$ 0.001) & 0.732 ($\pm$ 0.000) \\ 
Neural Network      & 0.755 ($\pm$ 0.030) & 0.782 ($\pm$ 0.006) & 0.736 ($\pm$ 0.001) \\
Random Forest       & 0.748 ($\pm$ 0.030) & 0.792 ($\pm$ 0.003) & 0.742 ($\pm$ 0.002) \\
Gradient Boosting   & 0.710 ($\pm$ 0.020) & \textbf{0.793} ($\pm$ 0.001) & \textbf{0.753} ($\pm$ 0.001) \\ \bottomrule
\end{tabular}
\caption{Mean (and std.) AUC scores in the testing dataset of models with best hyperparameters.}
\label{tab:credit_models_results}
\end{table}

\section{Fairness}
\label{sec:fairness}

Despite the success of machine learning in supporting decision-making for credit, debates about the potential of these algorithms to perpetuate discriminatory practices have increased considerably.  Such a concern is not unreasonable due to the growing number of examples where data-based methods are unintentionally "encoding" human biases (e.g. \citep{sweeney2013discrimination}, \citep{bolukbasi2019man}, \citep{caliskan2017semantics}, \citep{buolamwini2018gender}, \citep{gillis2021input}).  One popular example is the COMPAS (Profilational Offender Management Profiling for Alternative Sanctions) software, used in the United States to decide on arrest before trial by measuring a person's risk of reoffending. Unfortunately, this software has proven to make unfavorable decisions mostly for African Americans~\citep{angwin2016machine}. In automatic credit scoring, regulations in many countries have already defined mandates against using legally protected characteristics, such as gender and ethnicity~\citep{iapp2023global, madiega2023artificial, roberts2021chinese}. Yet, machine learning models can learn to replicate these biases from their data, even when the sensitive information is not present as a feature.

The topic of fairness in artificial intelligence ceased to be a niche and became a relevant field, even with a dedicated conference (ACM FAccT). Several surveys have explored machine learning fairness~\citep{mitchell2021algorithmic, mehrabi2021survey, pessach2022review, orphanou2022mitigating, wan2023inprocessing, shahbazi2023representation}, to which we refer the reader interested in learning more. In particular, \citet{kozodoi2022fairness} overviews fairness in the context of credit scoring. 

The diverse techniques already proposed to mitigate unfairness in machine learning can be applied at various stages within the conventional machine learning pipeline. Pre-processing mitigation methods are transformations applied in the data preparation stage, independent of any model~\citep{nielsen2020practical}. Following, in-processing methods alter the training procedures of the models to minimize not only the traditional loss but also a measure of fairness. This can be done through regularization~\citep{kamishima2012fairness}, reweighing samples~\citep{calders2009building, kamiran2012data}, adding constraints~\citep{zafar2017fairness, zafar2017fairnessbeyond, woodworth2017learning, cruz2023fairgbm}, multi-objective approaches~\citep{liu2022accuracy, padh2020adressing, zhang2021fairer, martinez2020minimax, zhang2023mitigating, guardieiro2023enforcing}, and others. Lastly, post-processors are applied only after training and do not need to change the model, working mainly with the threshold used for positive and negative classifications~\citep{hardt2016equality}.

In the following sections, we detail the definitions and metrics of fairness and present five mitigation methods: \textit{Reweighing}, \textit{Equal Opportunity and Demographic Parity Classifiers}, \textit{FairGBM}, and \textit{Threshold Optimizer}. These techniques will serve as good examples of the trade-offs of the different approaches and will be compared in our experiments.

\subsection{Definition and metrics}

One initial difficulty in developing fair algorithms is the definition of a fair decision and how to measure fairness. Different definitions and metrics have already been proposed, and most techniques for fairness in ML are designed to work with a subset of those metrics. Thus, a presentation and discussion of definitions and metrics of fairness is necessary. A common way to organize the fairness definitions, as \citet{mehrabi2021survey} presents, is to consider the discrimination against the individual or the collective. \textit{Individual fairness} dictates that similar people should have similar outcomes, while \textit{group fairness} says that different groups should have, on average, similar outcomes. \textit{Subgroup fairness} is a middle ground between these two that considers an increased quantity of groups defined based on the intersection of sensitive attributes~\citep{kearns2018preventing}.

\subsubsection{Group Fairness Metrics}

The notion of group fairness is the most common, also being considered by fairness regulations~\citep{wachter2021fairness}. 
It is based on restraining the relation between the algorithm outcome and a sensitive categorical attribute that separates the population into different groups.
Three different principles are considered to measure group fairness~\citep{varshney2022trustworthy,hertweck2022gradual}: \textit{independence} dictates that the model's predictions must be independent of the sensitive attribute; \textit{separation} asserts the subset of the data that obtained a favorable decision should not present any bias; and \textit{sufficiency} says that given the model prediction, the probability of the ground truth being positive must be the same for each group. A formulation of these concepts in terms of mutual information can be found in \citet{hertweck2022gradual}.

Applying these principles in credit scoring has different impacts. Models that satisfy independence will predict the same rate of accepted customers for each group, and according to \citet{barocas2023fairness}, the ability to repay a loan tends to be different in each group. Using the same probability to give loans to individuals of different groups can lead to more defaults in the unprivileged group. 
\citet{hardt2016equality} state that granting loans to individuals in unprivileged groups tends to worsen their financial situations in the long term.  Otherwise, \textit{separation} asks for the same error rates between the groups but allows different rates of positive classification, so a fair model can make more realistic credit allocation decisions. The formulation of the separation criterion is based on the example of credit scoring and the limitations of independence~\citep{hardt2016equality}. The decision of which equity constraint is best suited for credit scoring is still in discussion, and each application needs to consider this with attention. It is crucial to study the long-term effects of the justice restrictions implemented to conclude whether such interventions have achieved the social objective of improving financial equality among different demographic groups.

We briefly remind from the defined notation, in which $X$ is the set of features and $Y$ is the target, with $Y=1$ if the client defaults and the beneficial outcome is $Y=0$. The model prediction is $\hat Y$ and sensitive variable $Z$  with values $1$ (unprivileged) and $0$ (privileged). Different metrics of group fairness are used to consider the three principles and some of them are defined as follows:

\begin{itemize}
    \item \textbf{Demographic Parity Difference (DPD):} is a measure of independence that is defined as the difference between the probabilities of privileged and unprivileged groups accessing the beneficial outcome: 
    \begin{equation}
        DPD = P(\hat{Y} = 0 \mid  Z=1)-P(\hat{Y} = 0 \mid  Z=0)
    \end{equation}
    \item \textbf{Equal Opportunity Difference (EOD):} is a measure of separation that is defined as the difference between the probabilities of privileged and unprivileged groups accessing the beneficial outcome when the individual deserves that outcome:
    \begin{equation}
      EOD =  P(\hat{Y}=0 \mid  Z=1,  Y=0) -P(\hat{Y}=0 \mid Z=0,  Y=0)  
    \end{equation}
    \item \textbf{Average Odds Difference (AOD):} is a measure of separation that considers true positive and true negative rates. It is defined as the average of two differences: the difference between the probabilities of beneficial outcomes when the individual deserves that outcome; and when the individual does not deserve that outcome. This metric extends the EOD by also considering the difference in the false positive rate.
    \begin{equation}
        \begin{split}
            \mathcal{D}_{i} = &P(\hat{Y}=0\mid  Z=1,  Y=i)  - P(\hat{Y}=0\mid   Z=0,  Y=i) \\
            AOD &= (\mathcal{D}_{0} + \mathcal{D}_{1}) / 2
        \end{split}
    \end{equation}
    \item \textbf{Average Predictive Value Difference (APVD):} is a metric considering the sufficiency principle and calibration. It is the mean of two differences: the difference between the probabilities of privileged and unprivileged groups that obtained the beneficial outcome to deserve this outcome; and the difference between the probabilities of privileged and unprivileged groups that were denied the benefit to deserve the beneficial outcome:
    \begin{equation}
        \begin{split}
            &\overline{\mathcal{D}}_i = P(Y=0 \mid  Z=1,  \hat{Y}= i) -P(Y=0 \mid   Z=0,  \hat{Y}=i) \\
            &APVD = (\overline{\mathcal{D}}_0 + \overline{\mathcal{D}}_1)/2
        \end{split}
    \end{equation}
\end{itemize}

In practice, the probabilities are approximated using the empirical distribution of samples.
With all metrics, the lower the value, the fairer the model is, with $0$ being the minimal value.

\subsubsection{Subgroup Fairness Metrics}

A common phenomenon when dealing with fairness is that, even though we can analyze different sensitive attributes separately and conclude that a model is fair, unfairness can be present in groups with more than one sensitive attribute. The definitions and metrics for subgroup fairness follow the ideas for group fairness, with the generalization of moving from binary groups to a larger quantity of groups. To deal with that, some approaches might measure how much each group differs from an average measure \citep{kearns2018preventing}.

\subsubsection{Individual Fairness Metrics}

A problem that arises when approximating metrics calculated from large groups is that individuals inside each group can end in an unfavorable condition in comparison to the average condition of the group~\citep{kearns2018preventing}. This raises the concern that to achieve fairness it is necessary to focus on making similar decisions to similar individuals. With this idea, one crucial detail is measuring the similarity between individuals. We present some measures of individual fairness based on \citet{varshney2022trustworthy}:

\begin{itemize}
    \item \textbf{Consistency:} For each individual, the consistency is 1 if the prediction is the same as the average prediction of the $k$ nearest neighbors. The consistency of the model is evaluated with the average consistency for all individuals. This metric depends on the definition of a distance to determine nearest neighbors, and different formulations arise, such as using or not using the sensitive attribute.
    \item \textbf{Counterfactual Fairness:} Consider the case where a man applies for a loan and receives a positive outcome. To have counterfactual fairness, a model should output the same prediction if the sensitive attribute is altered and the remaining attributes are kept the same.
\end{itemize}

Despite the importance of individual considerations in fairness, most research and regulations focus on group fairness. Therefore, our study will not consider techniques designed for individual fairness.

\subsection{Methods}

\subsubsection{Pre-processing}

As previously mentioned, pre-processing is the bias mitigation method applied at the initial stage. Some advantages of pre-processing a dataset are that the new data can be used for any downstream task, and it is unnecessary to modify the model. However, this method is inferior in performance in terms of precision and fairness. Furthermore, pre-processing can only be used to optimize a limited number of fairness metrics, as we do not have the label information at this stage. 

\myparagraph{Reweighing.} Proposed by \citet{calders2009building}, consists of adding weights to each training data sample according to the sensitive attribute. Therefore, it can only be applied by models incorporating weights in their learning.
Designed for classification tasks and categorical sensitive attributes, the weights are defined to enforce the labels to be independent of the sensitive attribute. If there is independence, we have that $P(Y = y \land Z = z) = P(Y=y) \times P(Z=z)$; however, most of the datasets will not satisfy this property. The weights are defined for each group $z$ and label $y$ as $W_{y,z} = \dfrac{P(Y=y) \times P(Z=z)}{P(Y=y \land Z=z)}$ and, in practice, the probabilities are calculated from the empirical distribution of the observations:

\begin{equation}
   W_{y,z} = \dfrac{\left(\dfrac{1}{n}\sum_{i=1}^{n} I_{[Y_i = y]}\right) \times \left(\dfrac{1}{n}\sum_{i=1}^{n} I_{[Z_i = z]}\right)}{\dfrac{1}{n}\sum_{i=1}^{n} I_{[Y_i = y \land Z_i = z]}} 
\end{equation}

$I_A$ is an indicator variable that is $1$ if $A$ is true and 0 otherwise. Consider the example where there is a 50\% percentage of men in the dataset, and 50\% of the labels for men are positive. However, only 20\% of the labels for women are positive. There is a total of 35\% of positive labels. The actual probability of a sample being woman and positive is 10\%. Still, the expected probability is $0.35 \cdot 0.5 = 17.5\%$, so the weight of a sample of a woman with positive label is $W_{1, \text{woman}} = 17.5/10 = 1.75$, while the weight of a sample a man with positive label is $W_{1, \text{man}} = 17.5/25 = 0.7$. The reweighing strategy updates the weights so that the not-so-well-observed scenarios (positive outcomes for the unprivileged group) are also significant in the loss value.

\subsubsection{In-processing}

Another category of bias mitigation methods is in-processing, which encompasses methods that directly modify the training algorithms. In other words, the decision-making process is modified by changing the feedback used to assess whether the decision was reasonable. Those methods allow flexibility in choosing which trade-off between fairness and precision is the most suited for the use case.
However, in-processing techniques are frequently designed for specific tasks, metrics, and models. They also tend to have higher computational costs than the pre- and post-processing methods.

\myparagraph{Demographic Parity/Equal Opportunity Classifier.} Logistic Regression is a simple yet very reliable model for credit scoring, as our experiments showed in Sec.~\ref{sec:credit_models}, and can be adjusted to incorporate fairness.  \citet{zafar2017fairness} designed an in-processing mitigation method that changes the loss function of convex margin-based methods, particularly logistic regression. This approach minimizes the prediction error with a constraint of max correlation between the sensitive attribute and the prediction, i.e., ensuring independence. Let $\theta$ be the vector of parameters of the logistic model, the constraint is formulated as:
\begin{equation}
  \left|\dfrac{1}{n}\sum_{i=1}^{n}(Z_i - \overline Z)\theta^TX_i\right| \leq c  
\end{equation}

Where $\overline Z$ is the mean of the sensitive attribute of samples. This constraint can be formulated to consider demographic parity, as presented above, or for equality of opportunity, by performing the sum only with the samples with positive labels. This technique is generally used by training with different values of $c$ and selecting a model from the results based on the trade-off between performance and fairness.

\myparagraph{FairGBM.} Gradient boosting is among the state of the art for tabular data, as our experiments at Sec.~\ref{sec:credit_models} depicted. Recently, different mitigation methods have already been used to incorporate fairness in this method~\citep{bhaskaruni2019improving, iosifidis2019adafair, grari2019fair, ravichandran2020fairxgboost, vargo2021individually}. In particular, FairGBM~\citep{cruz2023fairgbm} is a modification of the gradient-boosting algorithm that enforces fairness with constrained learning. A fairness metric is also minimized jointly with the prediction error at each iteration. One important detail is that, as the boosting algorithm minimizes differentiable functions, it was necessary to use a differentiable proxy function of the fairness metric, which they developed for demographic parity and equality of opportunity.
 
\subsubsection{Post-processing}

Post-processing methods are widely used when facing a “black box” model. They alter the model output to satisfy a particular fairness constraint, such as defining separated thresholds of positive prediction for each group. Despite their versatility, they do not have the flexibility to choose the trade-off between accuracy and fairness and can produce worse improvements than other approaches.

\myparagraph{Threshold Optimizer.} Proposed by \citet{hardt2016equality}, this derives a separated threshold for each group by utilizing the ROC curve. The ROC curve is commonly created by calculating the false positive and true positive rates for different threshold values. But, in this mitigation method, the curve is calculated for each group, i.e., the rates are calculated for the subset of samples of each group. If the original score is fair, the curves will be similar, and the threshold could be the same for the groups. However, this is not generally the case. Then, a point of minimal prediction loss is searched at the intersection of the convex hull of ROC curves. The prediction loss can be minimized inside this region without unfairness because the searched area is of fair solutions, and the selected point will have different threshold values for each group.

\subsection{Fairness Experiments}

An experiment was performed to compare the different unfairness mitigation methods presented. \giovani{We utilized the client's gender as the protected attribute, separating the population into two groups. Baseline algorithms were trained with and without the protected attribute for comparison. Pre-processing and in-processing utilized the protected attribute for optimization; however, this feature is not necessary during inference (the feature is not an input of the model obtained by the algorithm). The post-processing algorithms utilized the protected attribute for training and inference, as it is a limitation of most post-processing solutions.}

The techniques have different training procedures, and minor modifications were implemented to accommodate each methodology. Hyperparameter tuning was performed to access the capabilities of the pre-processing and in-processing techniques. Reweighing was applied to the data before training, and the sample weights obtained were used to train all the models from Sec.~\ref{sec:credit_models}, performing hyperparameter optimization for the standard parameters of each classifier. Hyperparameter optimization of in-processing algorithms considered parameters related to model complexity and fairness strength. The threshold optimization method was applied to the models obtained in Sec.~\ref{sec:credit_models} and does not need hyperparameter tuning.
\giovani{For pre-processing and in-processing techniques the threshold for classification was selected to maximize the Kolmogorov–Smirnov statistic of the distribution of class scores}.

It is essential to consider that models should not focus only on maximizing performance, as this could prevent them from achieving fairness. We formulate the following score to maximize with hyperparameter optimization: (\texttt{perf}) $- (\mathbb{I}_{[\textrm{\texttt{fair}} \leq f^\star]}) (M |\textrm{\texttt{fair}} - f^\star|)$. This metric has a fairness goal ($f^\star$), and when this goal is reached, the model will be evaluated regarding only its performance (\texttt{perf}). When the fairness metric of the model \texttt{ fair} is not lower than $f^\star$, a high value $M$ penalizes the score with the distance from the fairness goal. \texttt{perf} should be a performance metric that is better when closer to $1$, and \texttt{fair} is a fairness metric that is better when closer to $0$. Our experiments were carried out with ROC as \texttt{perf}, EOD as \texttt{fair}, and the objective of fairness was $f^\star = 0.05$. Due to the results of the baseline models, as we discuss below, we used $f^\star = 0.01$ with the Taiwan dataset as the baseline algorithms already reached an EOD lower than 0.05.  

\begin{table}[!t]
\centering
\resizebox{0.9\linewidth}{!}
{
\begin{tabular}{lccccc}
\hline
Model               & Bal. Acc.              & EOD                    & DPD                    & APVD                    \\ 
\hline
\multicolumn{5}{c}{German}                                                                                               \\ 
\hline
LR Unaware & \textbf{0.701 ± 0.014}          & 0.073 ± 0.037          & \textbf{0.041 ± 0.03}          & 0.094 ± 0.028           \\ 
\hline
NN Unaware      & 0.696 ± 0.033          & 0.066 ± 0.058          & 0.086 ± 0.047          & \textbf{0.088 ± 0.025}           \\ 
\hline
RF Unaware       & 0.688 ± 0.025          & 0.120 ± 0.056          & 0.067 ± 0.044          & 0.112 ± 0.018           \\ 
\hline
GBM Unaware   & 0.656 ± 0.033          & \textbf{0.041 ± 0.035}          & 0.048 ± 0.025          & 0.103 ± 0.041           \\ 
\hline
LR Aware & 0.697 ± 0.024          & 0.193 ± 0.060          & 0.124 ± 0.055          & 0.137 ± 0.026           \\ 
\hline
NN Aware      & 0.672 ± 0.049          & 0.177 ± 0.169          & 0.145 ± 0.147          & 0.132 ± 0.040           \\ 
\hline
RF Aware       & 0.672 ± 0.027          & 0.181 ± 0.070          & 0.103 ± 0.065          & 0.132 ± 0.024           \\ 
\hline
GBM Aware   & 0.632 ± 0.041          & 0.090 ± 0.060          & 0.066 ± 0.037          & 0.130 ± 0.031           \\ 
\hline
\multicolumn{5}{c}{Taiwan}                                                                                               \\ 
\hline
LR Unaware & 0.709 ± 0.001          & \textbf{0.018 ± 0.002}          & 0.044 ± 0.002          & 0.040 ± 0.001           \\ 
\hline
NN Unaware      & 0.711 ± 0.004          & 0.026 ± 0.005          & 0.050 ± 0.004          & 0.035 ± 0.003  \\ 
\hline
RF Unaware       & \textbf{0.720 ± 0.003} & 0.020 ± 0.006          & \textbf{0.041 ± 0.006}          & 0.037 ± 0.003           \\ 
\hline
GBM Unaware   & 0.721 ± 0.004 & 0.023 ± 0.006          & 0.045 ± 0.006          & 0.036 ± 0.002           \\ 
\hline
LR Aware & 0.709 ± 0.001          & 0.051 ± 0.020          & 0.078 ± 0.019          & 0.022 ± 0.009           \\ 
\hline
NN Aware      & 0.713 ± 0.003          & 0.074 ± 0.016          & 0.099 ± 0.015          & \textbf{0.013 ± 0.005}  \\ 
\hline
RF Aware       & 0.721 ± 0.004 & 0.032 ± 0.007          & 0.052 ± 0.005          & 0.033 ± 0.002           \\ 
\hline
GBM Aware   & 0.721 ± 0.002 & 0.049 ± 0.007          & 0.072 ± 0.007          & 0.025 ± 0.003           \\
\hline
\multicolumn{5}{c}{Home Credit}                                                                                           \\ 
\hline
LR Unaware & 0.671 ± 0.001          & \textbf{0.036 ± 0.001}          & \textbf{0.045 ± 0.001}          & 0.028 ± 0.000  \\ 
\hline
NN Unaware      & 0.674 ± 0.001          & 0.043 ± 0.002          & 0.053 ± 0.002          & 0.027 ± 0.000  \\ 
\hline
RF Unaware       & 0.677 ± 0.001          & 0.077 ± 0.002          & 0.086 ± 0.002          & 0.022 ± 0.000           \\ 
\hline
GBM Unaware   & \textbf{0.687 ± 0.001} & 0.070 ± 0.003          & 0.080 ± 0.003          & 0.023 ± 0.001           \\ 
\hline
LR Aware & 0.675 ± 0.000          & 0.174 ± 0.003          & 0.182 ± 0.002          & \textbf{0.010 ± 0.000}  \\ 
\hline
NN Aware      & 0.679 ± 0.001          & 0.168 ± 0.007          & 0.178 ± 0.007          & \textbf{0.010 ± 0.001}  \\ 
\hline
RF Aware       & 0.679 ± 0.002          & 0.153 ± 0.011          & 0.162 ± 0.012          & 0.012 ± 0.002           \\ 
\hline
GBM Aware   & 0.688 ± 0.002 & 0.153 ± 0.003          & 0.162 ± 0.003          & 0.011 ± 0.000           \\ 
\hline
\end{tabular}
}
\caption{Average performance and fairness of baseline models on three datasets. ``Unaware'' indicates that the gender feature was not utilized by the model, while ``Aware'' indicates that it was.}
\label{tab:fairness_baselines}
\end{table}

\giovani{Results at Tab.~\ref{tab:fairness_baselines} presents a comparison of baseline classifiers without (unaware) and with (aware) access to the sensitive attribute as a variable. Results are separated by each dataset and include the average performance and fairness. We display balanced accuracy as it is impossible to calculate ROC for models with Threshold Optimizer, as they do not have a probability prediction. We also opt not to present AOD results due to their similarities with EOD. The algorithms presented did not had their hyperparameters optimized considering fairness, but only to maximize the ROC score. In all data sets, the unaware algorithms obtained the highest performance and lowest EOD (the lower the better). The decrease in EOD was significant between unaware and aware classifiers without a cost in performance. For example, the best EOD in the German dataset was 0.041 between the unaware classifiers and 0.090 between the aware classifiers. This was similar for the DPD metric. Aware classifiers obtained better values with the APVD metric, indicating that they are more fairly calibrated, even though they do not generate fair predictions. The results on the Taiwan dataset show a very interesting property, as almost all models had EOD lower than $0.05$. This shows that the models' fairness is also related to the dataset utilized. The results show that removing the sensitive attribute already presented a reduction in bias, which can be further improved utilizing fairness techniques, as we discuss in the following.}

\giovani{
Results for fair algorithms are presented at Tab.~\ref{tab:fairness_results}. We present the average results at the 10-folds, including the same metrics as the ones used previously. We added two new columns: the difference in balanced accuracy and EOD of each model compared to a baseline. The pre-processing and post-processing methods are compared to the original model. FairGBM is compared with LGBM. Demographic Parity and Equal Opportunity Classifiers are compared against Logistic Regression. The result for the Home Credit dataset using Equal Opportunity/Demographic Parity Classifiers is not present as the optimization of these algorithms caused the system to run out of memory. 

In all data sets, the Threshold Optimizer approach reached the fairness goal and presented the lowest EOD, with respective values of $0.035$, $0.005$, and $0.002$. It presented a decrease in balanced accuracy; however, it was lower than $10^-2$ and was the best balanced accuracy in Taiwan. Although these results are positive, Threshold Optimizer has the downside that it requires sensitive attributes to make predictions. Consequently, the model would need access to the gender information of all future clients, which may be legally prohibited. Considering the remaining methods, the best EOD was obtained by Reweighting with gradient boost in German, Equal Opportunity Classifier in Taiwan, and Reweighting with logistic regression in Home Credit, which were also able to reach the fairness goal. Almost all techniques presented a decrease in EOD (the lower the better) accompanied by a decrease in balanced accuracy (the higher the better), highlighting the trade-off between two metrics. With the exception of the German dataset, the baseline models obtained better values on APVD than the fair approaches. This occurs because current techniques do not focus on the calibration of models.
}

\begin{table}[!t]
\centering
\resizebox{\linewidth}{!}
{
\begin{tabular}{lccccccc} 
\hline
Model               & Diff in Perf. & Diff in Fair. & Bal. Acc.     & EOD                    & DPD                    & APVD                    \\ 
\hline
\multicolumn{7}{c}{German}                                                                                               \\ 
\hline
Reweighing LR   &  -0.012 ± 0.038 &  0.023 ± 0.064   &  0.689 ± 0.033 & 0.097 ± 0.049 & 0.057 ± 0.039 & 0.092 ± 0.018   \\
\hline
Reweighing NN   &  -0.036 ± 0.07  & 0.006 ± 0.083    & 0.661 ± 0.068  & 0.071 ± 0.044 & 0.079 ± 0.055  & 0.080 ± 0.029   \\
\hline
Reweighing RF  & -0.009 ± 0.038 & -0.032 ± 0.083  &   0.678 ± 0.027 & 0.088 ± 0.048 & 0.045 ± 0.036 &  0.095 ± 0.020\\
\hline
Reweighing GBM &  -0.035 ± 0.047 & -0.001 ± 0.059  &  0.620 ± 0.069  &  0.040 ± 0.041 &  \textbf{0.038 ± 0.040} & 0.123 ± 0.041         \\ 
\hline
DP Classifier & 0.009 ± 0.020 & -0.023 ± 0.050   &   \textbf{0.710 ± 0.014}  & 0.050 ± 0.032  & 0.061 ± 0.050 & \textbf{0.065 ± 0.024}\\ 
\hline
EO Classifier &  0.007 ± 0.018 &  -0.020 ± 0.058  &    0.708 ± 0.020 & 0.053 ± 0.042 & 0.043 ± 0.041 &  0.074 ± 0.020 \\ 
\hline
FairGBM  & -0.037 ± 0.079  & 0.016 ± 0.072   &  0.618 ± 0.075 & 0.057 ± 0.067 & 0.052 ± 0.038 & 0.129 ± 0.078\\ 
\hline
Thr. Opt. LR & -0.006 ± 0.022 & 0.045 ± 0.052  &   0.695 ± 0.016 & 0.118 ± 0.039 & 0.059 ± 0.034 & 0.105 ± 0.011\\ 
\hline
Thr. Opt. NN & -0.003 ± 0.020 &  0.029 ± 0.028   &   0.694 ± 0.030 & 0.095 ± 0.061 & 0.088 ± 0.066 & 0.096 ± 0.031 \\
\hline
Thr. Opt. RF & -0.001 ± 0.015  & -0.03 ± 0.043   &  0.686 ± 0.024  & 0.090 ± 0.044 & 0.039 ± 0.025 & 0.107 ± 0.016\\ 
\hline
Thr. Opt. GBM & 0.002 ± 0.011 & -0.006 ± 0.031   &  0.658 ± 0.026 & \textbf{0.035 ± 0.031} & 0.054 ± 0.036  & 0.090 ± 0.021\\ 
\hline
\multicolumn{7}{c}{Taiwan}                                                                                               \\ 
\hline
Reweighing LR & -0.001 ± 0.004  &   0.000 ± 0.003  &   0.708 ± 0.005 &  0.018 ± 0.002 & 0.044 ± 0.002  & 0.040 ± 0.002\\ 
\hline
Reweighing NN  &  -0.087 ± 0.107  & -0.010 ± 0.016  &   0.624 ± 0.107  & 0.015 ± 0.014 & 0.031 ± 0.027 & \textbf{0.036 ± 0.004} \\ 
\hline
Reweighing RF & -0.003 ± 0.006 &   -0.005 ± 0.006 &   0.717 ± 0.005 &   0.016 ± 0.005 &   0.037 ± 0.002 &   0.039 ± 0.003  \\ 
\hline
Reweighing GBM & -0.036 ± 0.067 &    -0.01 ± 0.011 &   0.685 ± 0.067 &   0.014 ± 0.009 &   0.033 ± 0.014 &   0.041 ± 0.005 \\ 
\hline
DP Classifier & -0.004 ± 0.004    & -0.01 ± 0.005   & 0.705 ± 0.003   & 0.008 ± 0.005    & 0.030 ± 0.009   & 0.049 ± 0.005 \\ 
\hline
EO Classifier  & -0.005 ± 0.005   & -0.009 ± 0.006   & 0.704 ± 0.005   & 0.008 ± 0.007   & 0.028 ± 0.011   & 0.049 ± 0.006 \\ 
\hline
FairGBM  & -0.015 ± 0.024   & -0.011 ± 0.012   & 0.706 ± 0.022   & 0.012 ± 0.007   & 0.035 ± 0.008   & 0.043 ± 0.005 \\ 
\hline
Thr. Opt. LR  & -0.0 ± 0.002   & -0.013 ± 0.004   & 0.709 ± 0.002   & \textbf{0.005 ± 0.003}   & 0.022 ± 0.003   & 0.052 ± 0.002 \\ 
\hline
Thr. Opt. NN & -0.002 ± 0.001    & -0.02 ± 0.008    & 0.710 ± 0.004   & 0.006 ± 0.003   & 0.022 ± 0.006    & 0.050 ± 0.004 \\
\hline
Thr. Opt. RF  & -0.001 ± 0.002   & -0.014 ± 0.011   & 0.719 ± 0.003   & 0.006 ± 0.005   & \textbf{0.016 ± 0.006}   & 0.048 ± 0.003 \\ 
\hline
Thr. Opt. GBM & -0.001 ± 0.001   & -0.015 ± 0.008    & \textbf{0.720 ± 0.005}   & 0.008 ± 0.004   & 0.015 ± 0.005   & 0.048 ± 0.003 \\ 
\hline
\multicolumn{7}{c}{Home Credit}                                                                                           \\ 
\hline
Reweighing LR & -0.001 ± 0.001    & -0.030 ± 0.002   & 0.671 ± 0.001   & 0.007 ± 0.001   & 0.015 ± 0.001     & 0.032 ± 0.000 \\ 
\hline
Reweighing NN & -0.001 ± 0.001   & -0.032 ± 0.004   & 0.673 ± 0.001   & 0.011 ± 0.003   & 0.021 ± 0.003   & 0.032 ± 0.001 \\ 
\hline
Reweighing RF  & -0.004 ± 0.008    & -0.03 ± 0.003   & 0.673 ± 0.007   & 0.047 ± 0.002   & 0.056 ± 0.002     & \textbf{0.027 ± 0.000} \\ 
\hline
Reweighing GBM & -0.001 ± 0.002   & -0.037 ± 0.004   & \textbf{0.686 ± 0.001}   & 0.033 ± 0.001   & 0.044 ± 0.001     & 0.029 ± 0.000 \\
\hline
FairGBM & -0.026 ± 0.01   & -0.029 ± 0.005    & 0.661 ± 0.010   & 0.041 ± 0.006    & 0.050 ± 0.006   & 0.029 ± 0.001 \\ 
\hline
Thr. Opt. LR & -0.002 ± 0.001   & -0.035 ± 0.002   & 0.669 ± 0.001   & \textbf{0.002 ± 0.001}   & 0.006 ± 0.001     & 0.033 ± 0.000 \\ 
\hline
Thr. Opt. NN & -0.003 ± 0.001   & -0.039 ± 0.002   & 0.672 ± 0.001   & 0.004 ± 0.002   & \textbf{0.005 ± 0.001}   & 0.034 ± 0.001 \\
\hline
Thr. Opt. RF  & -0.002 ± 0.001   & -0.075 ± 0.002   & 0.675 ± 0.001   & \textbf{0.002 ± 0.001}   & 0.007 ± 0.001     & 0.033 ± 0.000 \\ 
\hline
Thr. Opt. GBM  & -0.003 ± 0.002   & -0.066 ± 0.005   & 0.684 ± 0.002   & 0.004 ± 0.002   & 0.006 ± 0.002   & 0.035 ± 0.001 \\ 
\hline
\end{tabular}
}
\caption{Average performance and fairness of all the presented techniques on three datasets.}
\label{tab:fairness_results}
\end{table}


It is challenging to select the ``best'' mitigation methodology, as performance varies depending on the dataset and context. The in-processing methods, like DP/EO Classifiers and FairGBM, are inherently tied to specific model types, which can limit their applicability if those models are not the top performers on a given dataset. We recommend prioritizing performance and addressing fairness through techniques like Reweighing and Threshold Optimizer, a sound strategy. The paper's results support this, showing that both methods improve fairness across various scenarios. However, Reweighing might face challenges with APVD (Average Predictive Value Difference) in some instances, such as the German dataset, indicating potential limitations in achieving specific fairness goals.

\section{Reject Inference}
\label{sec:reject_inference}

In machine learning systems for credit scoring, historical data plays a pivotal role in constructing and updating the classifier. However, this data can sometimes fail to comprehensively represent the entire population, leading to unfair distortions for specific groups, as we discussed in the previous section. Furthermore, a secondary source of bias in the data emerges when the data used comprises only previously accepted clients, known as sample bias ~\cite{li2017reject}. Consequently, to predict the likelihood of default of a new client, the model will only consider the distribution of the accepted clients, and with each iteration of this process, the classification becomes increasingly ill-suited to categorize new data that diverges from the distribution of approved clients~\cite{song2022towards}. As the reject decision is based on the client attributes, the missing data is not random, and the observed and unobserved populations will have different distributions.

Some techniques, known as \textit{Reject Inference} (RI), attempt to overcome this problem by incorporating the unlabeled data from rejected clients in various ways. The most basic approach is to assume all rejected clients are bad payers, which can be problematic, as it considers that the classification is perfect. Augmentation is a more intricate RI technique that weights the importance of accepted clients by their proximity to the distribution of the rejected population. Extrapolation techniques try to infer the label of the rejected clients based on the distribution of the accepted clients~\cite{li2017reject, kozodoi2019shallow, liao2021data, kang2021graph, annas2023semi}. However, a more assertive, and expensive, strategy would be to collect more data by accepting clients that would be rejected or borrowing data from other sources/companies. In the era of deep learning, techniques are making use of semi-supervised learning approaches that use the labeled data to train a model and propagate this prediction to unlabeled samples~\cite{kozodoi2019shallow, liao2021data, kang2021graph, annas2023semi}. Lastly, outlier detection algorithms can also be used to find outliers inside the rejected population and change their label to good payers~\cite{xia2019novel}. 

The usefulness of reject inference have already been confirmed by many studies that identified improved performance when using these techniques~\cite{li2017reject, mancisidor2020deep, shen2020three}. Furthermore, reject inference strategies can bring more than a positive economic impact. These techniques are also capable of a positive social impact on historically marginalized populations by credit scoring systems and should be considered even with a loss of performance. Following, we define the problem of reject inference and present a selection of methods. Finally, we discuss the results of our experiments that evaluated the presented methods.

\subsection{Definition of Reject Inference}

\begin{figure}[htbp]
    \centering
    \includegraphics[width = \linewidth]{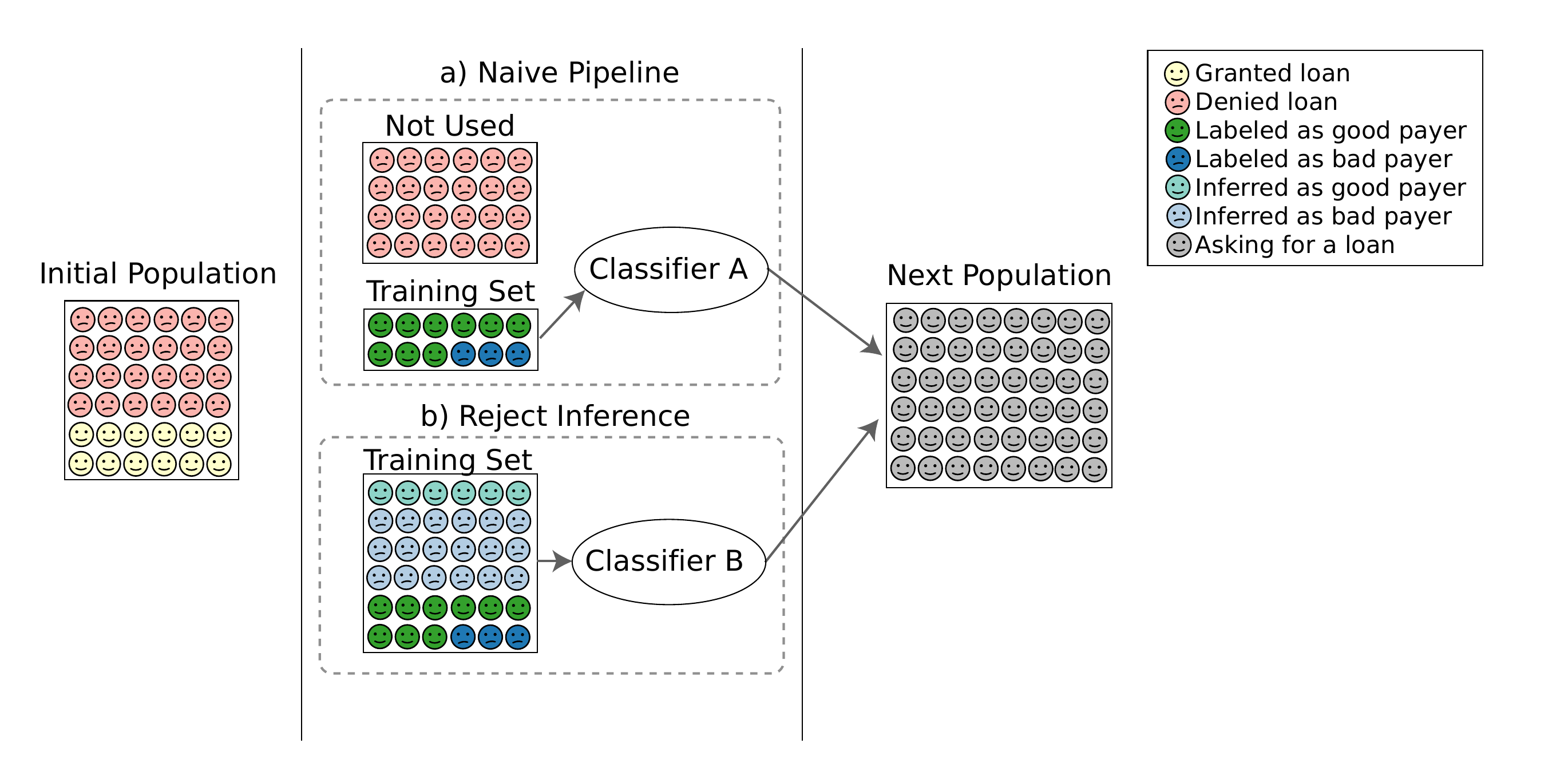}
    \caption{Naive pipeline a) versus pipeline that uses reject inference b). The credit was not approved for the red population on a previous classification iteration. Thus, we only have labels for the yellow population. The red clients have their samples a) discarded from the training dataset and b) inferred (clear green and blue). Thus, Classifier A will classify the next population of clients not knowing the rejected population, while Classifier B will classify the next population, with more robust results on the rejected population.}
    \label{fig:pipelineRI}
\end{figure}

Reject inference is more commonly known as a set of techniques that try to infer if a client who already had their credit request denied would fit in the class of \textit{good} or \textit{bad} payers~\cite{shen2020three}. Fig.~\ref{fig:pipelineRI} illustrates the differences between a naive classification system for credit and one that uses reject inference. In the naive pipeline Fig.~\ref{fig:pipelineRI}~(a), several clients ask for a loan from a company that classifies them as either potential good or potential bad payers, regarding their attributes of interest. The accepted clients that took the loan are later labeled as good or bad payers, according to the criteria defined by the company that evaluates their paying behavior, for example, the number of months that the client is late for paying back their loan. The company then uses the data of these clients and their associated payment behavior to update its classification algorithm. Since the rejected clients can not have their payment behavior analyzed, their data is discarded in this scenario~\cite{li2017reject, mancisidor2020deep}.  Fig.~\ref{fig:pipelineRI}~(b) illustrates a scenario with a method of reject inference. The beginning of the classification process is similar to the typical one. However, instead of discarding the data of those rejected clients, some technique is applied to infer a label to these clients \cite{anderson2022credit}. After that, a portion of the data of the rejected clients and their inferred labels can be added to the new training data set, which is used to update the previous model. 

\subsection{Reject Inference Metrics}

Since the aim of reject inference is to better classify a population that lacks historical credit data, evaluating the true performance of a reject inference technique is quite challenging. Therefore evaluating a model with a test set that consists only of accepted samples \citep{kozodoi2019shallow},  will not represent the actual performance of the model in the whole population (with both accepted and rejected samples). To tackle this evaluation bias, \citet{annas2023semi} compares the performance of RI models with two types of test sets to evaluate model predictions: a biased test sample containing only accepted clients and an unbiased test set constructed with both accepted clients and rejected clients with inferred labels. Their results show a disparity in the results achieved by the studied models for each set. In our study, we consider the following two metrics:

\myparagraph{Kickout Metric.} Proposed by \citet{kozodoi2019shallow}, the \textit{kickout} metric aims to compare the set of applications accepted by a scoring model before and after employing a reject inference technique. It calculates the number of good and bad payers that were \textit{kicked out} from the initial set of accepted clients. As it is expected that a good technique would \textit{kickout} more bad payers, the higher the \textit{kickout} value, the better the performance of the model. The rejected samples are required for this metric, but not their true labels. The metric values range from -1 to 1, with values close to $1$ indicating that more bad payers were kicked out and values close to $-1$ indicating that more good payers were kicked out.

\myparagraph{Approval Rate.} \citet{liao2021data} introduce \textit{approval rate} as a new metric on the credit setting. This metric considers that a banking institution has a maximum of bad payers it is willing to accept. This metric first calculates a threshold $\tau$ that guarantees this acceptance rate of bad payers. Then, the \textit{approval rate} is the percentage of clients (from the complete population) whose risk score is under the selected threshold. The reasoning behind this metric is that a higher number of approved cases implies greater profitability for the company.

\subsection{Methods}

\subsubsection{Augmentation}

Augmentation is a strategy that adjusts the weights of accepted applicants in a way to represent the rejected population \citep{banasik2007reject}. Despite some works that argue that such an approach is limited~\citep{anderson2022credit}, it was one of the most common strategies used in credit. \citet{anderson2022credit} divides this strategy into two categories: upward and downward. On the upward approach, the weights of the accepted clients are updated to be indirectly proportional to the chances of a client being accepted. On the downward approach, the weights of the accepted clients are updated to be directly proportional to the chances of a client being rejected. 

However, other variations of this technique have already been proposed. \citet{siddiqi2012credit} cites augmentation with soft cut-off as a strategy that adjusts the weights of accepted samples based on a score of the group that the sample was placed. The score is based on the probability of a sample being approved from a model that approximates the initial step of approval/rejection. For each group, the augmentation factor is inversely proportional to the probability of a sample in that group being from an accepted client. \citet{anderson2022credit} also cite a variation of augmentation, called fuzzy-parcelling. In this technique, the rejected samples are duplicated, with half being labeled as bad and the other half being labeled as good payers. The half that was labeled as good receives as weight the probability of each sample being accepted and the other half gets as weight the probability of being rejected. The probability of being accepted and rejected comes from a model similar to the one used in the soft cut-off technique. The weight of the accepted samples is one. 

\subsubsection{Extrapolation}

We call extrapolation the strategy in which the performance of the accepted clients is used to infer the behavior of the rejected ones \cite{banasik2007reject} and is currently the most utilized technique in credit scoring~\cite{anderson2022credit}. \citet{siddiqi2012credit} refers to this technique as a hard cutoff and a simple augmentation. Some authors also sometimes call parceling. This is a pretty simple strategy in which a credit scoring model built on the accepted population is trained to classify good and bad payers. This model is used on the rejected samples to infer their labels. The rejected samples with their new inferred labels are then combined with the accepted population and used to train a new model.

The main problem of using simple extrapolation lies in the assumption that the population of rejected clients would behave exactly like the population of the accepted clients and have the same risk of defaulting \cite{anderson2022credit}. Because of this, \citet{banasik2007reject} considers this strategy, when applied with this assumption, useless but also harmless. However, \citet{anderson2022credit} states that this strategy is rarely used in isolation and that some techniques can enhance the usefulness of the extrapolated predicted probabilities.

\subsubsection{Label Spreading}

Label Spreading~\citep{zhou2003learning} is a semi-supervised method that was employed by \citet{kang2021graph} together with an augmentation technique to build a reject inference strategy. Label Propagation is a graph-based algorithm based on the assumption that nearby samples are likely to have the same labels. It is an algorithm that has been used not only in reject inference but also in many other problems revolving around unlabeled data. 

For the Label Spreading model, rejected samples are concatenated with the accepted samples. The rejected samples get $(-1)$ as a temporary label. The Label Spreading is then fitted with this new dataset. During the training, the rejected samples are then assigned new labels, according to their localization on the graph. These labels can then be retrieved and used to train other models \cite{kang2021graph}.

\subsection{Experiments}


To demonstrate and evaluate reject inference, we used data from the Home Credit dataset~\cite{homecredit}, previously present at Sec.~\ref{sec:datasets}. To simulate employing reject inference techniques, it is necessary to have a group of accepted and rejected clients, which is not present in the Home Credit dataset. We simulated a rejection policy to have both groups and applied it to separate the dataset samples based on the work of \citet{liu2023rmtnet}. First, we separated the features into two groups, one for model training and study of reject inference techniques and the other for building an accept/reject model. The samples from Home Credit were also divided into a small group for training the accept/reject model and the remaining data was divided into accepts and rejects using the previously trained model. Using the classifier probabilities of defaulting and a threshold of $0.4$, the reject group had $72\%$ of the samples, with a default rate of $9\%$. The accept group had $28\%$ of the samples, with a default rate of only $5\%$. The resulting proportions between accepts and rejects follow the proportions commonly found in credit datasets.

We selected LightGBM as our base classifier to test all the techniques. LightGBM was used to build our Accept/Reject models and our models for probability of default, including the baseline model (BM) built with only accepted samples. We selected the following reject inference augmentation techniques to be compared: Augmentation with Soft Cuttoff (SCA), Upward Augmentation (UA), Downward Augmentation (DA), and Fuzzy Augmentation (FA). We also chose 3 different variations of the Extrapolation technique: Only adding rejected samples inferred as bad payers to the new training set (BE), adding all rejected samples to the new training set (AE), and finally only adding the samples with the highest probability of being good payers and the ones with the highest probability of being bad payers (CE). We also selected a Label Spreading strategy (LS) for comparison. 

\begin{table}[th]
\centering
\begin{tabular}{lcccc}
\toprule
Model & AUC            & Bal. Acc.          & AR             & KK             \\ \hline
BM    & \textbf{0.739} & \textbf{0.677} & \textbf{0.859} & 0.000          \\ \hline
SCA   & 0.734          & 0.673          & 0.856          & 0.046          \\
DA    & 0.738          & \textbf{0.677} & 0.857          & 0.025          \\
UA    & 0.738          & 0.676          & 0.856          & 0.018          \\
FA    & 0.728          & 0.655          & 0.841          & 0.010          \\ \hline
AE    & 0.733          & 0.670          & 0.851          & 0.013          \\
CE    & 0.738          & 0.634          & 0.858          & 0.016          \\
BE    & 0.722          & 0.652          & 0.819          & 0.046          \\ \hline
LS    & 0.731          & 0.655          & 0.854          & \textbf{0.130}  \\ \bottomrule
\end{tabular}
\caption{Average metrics of performance for baseline model and reject inference techniques on the Home Credit dataset.}
\label{ri_table}
\end{table}


Each experiment was executed 5 times with a different random seed and the average results are displayed in Tab.~\ref{ri_table}. The table shows the results of the AUC score, balanced accuracy (using a threshold of $0.5$), approval rate (AR) and kickout (KK). For the AR metric, the threshold was calculated using a validation set for each model. For KK, a value of $0.15$ was utilized as approval rate.

Even though the baseline model had the highest AUC value, the models with reject inference did not presented a significant loss in performance. Upward and Downward Augmentation were the techniques that yielded the best AUC results among the augmentation techniques. The version of extrapolation that included only the most confident predictions for updating the training set achieved the best results among the extrapolation techniques for the AUC metric.

Considering the metrics specifics for reject inference, the baseline model had the best AR, while the extrapolation model that added only bad samples to update the training set showed the worst performance in this metric. This is not surprising, considering that this model alters the proportion of bad samples
Remaining techniques had similar values of AR. The Label Spreading technique obtained the highest KK metric. This indicates that the LS technique removed a bad payers from the final accepted samples, which was not obtained by the other techniques. 

If we analyze only the AUC and KK metrics we might conclude that reject inference techniques are not particularly effective at improving model performance. However, this interpretation overlooks reject role in addressing the inherent biases in classification tasks involving rejected samples. Recent literature indicates that more advanced reject inference techniques are outperforming baseline models~\cite{kang2021graph, kozodoi2019shallow, liao2021data}. The KK result for Label Spreading, the most recent strategy among those selected, supports this.


\section{Explainability}
\label{sec:explainability}

Trust remains a barrier in the use of algorithms for decision-making in social domains, and one of the major concerns is the lack of clarity in the rationale behind the decisions made. As these algorithms become increasingly complex, comprehending the decision process becomes more difficult. In this context, explainability techniques capable of extracting information from algorithm decisions have been more utilized~\cite{murdoch2019definitions}. In the context of credit scoring, governments have been formulating regulations~\cite{goodman2018european} that require knowledge about the rationale of the algorithms and systems employed that explain the decisions to individuals. Such regulations stimulated interpretable models, such as logistic regression, to remain the industry standard despite the performance advantages of more complex models~\cite{dastile2020statistical}.

With the diffusion of intelligent systems, the literature on explainability techniques is vast~\cite{molnar2022interpretable}. At a high level, techniques can be divided if they provide global or local explanations. Global explanations~\cite{friedman2001greedy, apley2020visualizing, friedman2008predictive, fisher2018all} address the general relation between the model output and its inputs, searching for which features are more relevant and if they have a positive or negative impact; while the local explanations are related to the relevant features for an individual decision~\cite{ribeiro2018anchors, ribeiro2018why, simonyan2014deep, smilkov2017smoothgrad, sundararajan2017axiomatic, lundberg2017unified, goldstein2015peeking}. Counterfactual explanations are a type of local explanation utilized in credit models, which tries to explain a decision by pointing out which values each feature should have to alter the outcome of the algorithm~\cite{watcher2017counterfactual, raimundo2024mining, verman2022counterfactual}. Lastly, a simpler approach frequently utilized in credit modeling is to restrain the complexity of models to improve interpretability, such as reducing the maximum depth of a decision tree or utilizing linear models (which can also have regularization to reduce complexity). A model with reduced complexity can have its decision described by simple relations from the features and can even use only a few of them to perform prediction. 

According to the regulation and intricacies of credit models, the use of explainability techniques in this context has two main objectives: model transparency and counterfactual explanations~\cite{bucker2022transparency}. The first objective covers the awareness of patterns learned by the algorithm, in contrast to the prior knowledge of domain experts. The second objective is crucial for individuals to act upon their situation so that a desirable outcome can be reached, such as the acceptance of a loan. With two very different characteristics, both types of explanations are of great relevance to the problem of credit scoring. In this section, we present a selection of explainability techniques related to both objectives and discuss their advantages. In the following, we exemplify some of the methods discussed to interpret models from Sec.~\ref{sec:credit_models}.

\subsection{Model Transparency}

When using automatic algorithms, practitioners can be unaware of the relevant features utilized in the decision process, which can lead to undesirable outcomes, such as biased outcomes or decisions based on spurious correlations. This poses the importance of model transparency, a set of techniques and analyses that are intended to bring into light the intricacies of the algorithm's decision process. Such techniques are also required by law on many countries, which states that decision-process with high impact on individual lives should be auditable. This section presents a selection of explainability techniques designed for model transparency, starting with global approaches such as interpretable models, partial dependence, and individual conditional expectation, and then two widely utilized local explainers, LIME and SHAP.

\textbf{Interpretable Models.} Certain models are considered interpretable because their decision-making process can be directly analyzed, mostly due to their reduced complexity. The most popular ones are logistic regression and small decision trees. For logistic regression, it is possible to interpret how the coefficients $\beta$ impact the log odds using the following relation obtained from the model formulation: $\ln \left(\frac{P(Y = 1| X_i)}{1 - P(Y = 1|X_i)}\right) = \beta^T X_i$.

 If the coefficient is positive for the $k$-th feature ($\beta_k > 0$), increasing $x_{ik}$ will increase the log-odds and consequently increase the probability of a positive outcome, while decreasing $x_{ik}$ will decrease it. On the other hand, if $\beta_k <0$, the relation is inverse. If all the features of $X$ are on the same scale, the magnitude of coefficient $|\beta_k|$ can be utilized to identify the most relevant features. An important detail is that this approach considers that changing one feature has no effect on the others; an assumption is generally invalid in real datasets.

 Despite using simple decision trees, gradient-boosting presents great complexity because of the large number of small estimators combined. However, it is also possible to obtain a measure of the importance of a feature from the importance of the feature during training~\cite{lange2022explainable}. As mentioned above (Sect.~\ref{sec:credit_models}), each node of a tree is based on a threshold of a feature that divides the samples into its child nodes. The training algorithm identifies the features and threshold values that better divide the samples according to the target. Counting how many times a feature is used to split the tree can be a method to measure the importance of the feature.

\textbf{Partial Dependence (PD)}~\cite{friedman2001greedy} works by analyzing what is the average model prediction for different values of a selected feature. To do that, it updates all the data points to have the same value of this feature, generates all the predictions, and calculates the average. By changing this fixed value, it is possible to draw a line plot of the feature value and the mean prediction.
Similar to interpreting the coefficients of a logistic model, this explanation assumes that the feature studied presents no correlation with other features, and changing the feature value can create unreal data points that could not be observed in real-world scenarios.

\textbf{Individual Conditional Expectation (ICE)}~\cite{goldstein2015peeking} is related to the PD technique and does the same process of changing a feature value for all samples but draws a line for each data sample. This can provide a more useful analysis, highlighting that a feature value affects and alters sample predictions in different ways. To produce a clearer figure, all lines can be positioned vertically to have their start at $0$, thus the vertical direction represents the difference in the prediction from the minimal value considered for the feature~\cite{molnar2022interpretable}. Yet, it still can be a cluttered visualization when many lines are present, and it also has the assumption of no correlation with the features that are not being analyzed.

\textbf{LIME} was proposed by~\citet{ribeiro2018why} and differently from previous approaches discussed, produces ``local'' explanations. The interpretation of the model behavior is only performed in the neighborhood of a particular data point. To obtain such explanations, it uses a perturbation-based approach: it generates new data points around the input point by making small perturbations in all feature values and generates the prediction of these points. The new points and predictions are then used to train an interpretable model, such as a linear regression, in which the coefficients can be used in the explanation. If a feature has a coefficient with a high absolute value, it is an important feature for the model in this data point.

\textbf{SHAP} is based on the work from Shapley~\cite{shapley1953contributions}, who presented a method to evaluate a player's contribution in the result of a game with the property that the sum of contributions is equal to the game reward. Despite this illustrative scenario, the same strategy can be used to attribute the contribution of feature values of a data point to the model prediction. \citet{lundberg2017unified} later presented a unified formulation of SHAP values for feature attribution. To obtain the features' attributions, it is necessary to calculate the difference in the prediction when the feature is present and when it is not. However, as most models are not capable of generating predictions with a reduced subset of features, this is measured by calculating the average prediction of a model with random values on features that are ``not used''. SHAP was initially designed for local explanations, however, the aggregation of feature attributions from multiple samples permits the interpretation of the model globally.

\subsection{Counterfactual Explanations}

 In some intelligent systems, the explanations provided to clients should guide them on how to change the obtained decision. In credit scoring, this type of explanation can be used by the client to alter his credit profile to achieve the desirable outcome, e.g., the acceptance of a loan. This justifies the development of counterfactual explanations~\cite{watcher2017counterfactual}, which are descriptions of small changes in feature values that are necessary to alter the model prediction. Explanations should be plausible, i.e., they should be achievable by a real client. This leads to different objectives when developing a counterfactual explanation~\cite{mothilal2020explaining}: they have proximity to the reference observation; they should alter a subset of features; they should be diverse so that the client can choose the more suited one. In this context, different approaches have been proposed to generate such explanations, and we exemplify them with one based on a search of feature space and one based on optimization.

\textbf{MAPOCAM} is a technique that avoids searching for all possible scenarios and generating the predictions. MAPOCAM~\cite{raimundo2024mining} is an algorithm that uses a tree-based search of counterfactual explanations that avoids performing unnecessary computations. It is a model-agnostic technique, but it has a lower computing cost if the model utilized has monotonic relations between features and the prediction. MAPOCAM also considers the different aspects of plausibility discussed above, with the option to search for counterfactuals that are the optimal solution for multiple objectives, such as the number of changes and the value of the highest change of feature values (on percentage).

\textbf{DiCE} was proposed by \citet{mothilal2020explaining} and through optimization, minimizes the distance of counterfactual explanations to the reference sample and maximizes the diversity among generated counterfactuals. Maximizing diversity enforces different explanations to utilize different sets of features. Thus, the client will have a more diverse set of options to act upon. The implementation permits the user to select how many counterfactuals are generated, which features can have their values changed, and trade-off parameters among sparsity and diversity.

\subsection{Data and Model Handling}

Explainability techniques are very diverse in the way which they interact with the data and with the model. This section describes important implementation details to obtain the best results when applying such techniques. As previously discussed in Sec.~\ref{sec:credit_models}, models are combined with a step of data pre-processing, which includes imputation of missing data, scaling of numerical features, and one hot encoding of categorical ones. The pre-processing needs to be considered when applying explainability techniques, to avoid, for example, calculating explanations for invalid values of categorical features. The following guidelines were implemented in the supplementary code. 

\begin{itemize}
    \item PDP and ICE: partial dependence plot can handle numeric or categorical features, while ICE can be generated only for numeric ones. The resulting plot should be presented in the original data scale to enrich the interpretation. This result is obtained by calculating the average prediction with inputs $x$ already scaled, and after that, the inverse operation is applied in $x$ but the average predictions are kept the same.
    \item SHAP: some implementations of SHAP utilize internal information of the model, while some are model-agnostic explainers.  In particular, the \textit{permutation explainer} properly handles categorical features, without treating them as numbers. While scaling of features does not alter SHAP values, the interpretation of the feature attributions is different when considering categorical features. Features attribution for categorical features can be obtained by summing the feature attribution of the new features created due to one hot encoding or by combining the encoding of categorical features and the model as only one block, which was done in this study.
    \item LIME: this approach will produce small perturbations around the observation to evaluate how such changes alter the prediction of the model. This perturbations need to be done at the same scale as the features and should be done differently for numeric and categorical features. LIME implementation can handle such details, being only necessary to inform which features are categorical and considering the scaling, one hot encoding, and model as one closed block.
    \item MAPOCAM: Despite being able to generate counterfactual explanations if categorical features are transformed into integers to represent the categories, it considers that such integers have an ordering to calculate a measure of distance to the reference individual. This is only correct if the categorical feature is binary. For that reason, the set of features that can be used in the counterfactual explanation should not include categorical features with more than two categories. More interpretable results are obtained by using the scaling, one hot encoding, and the model as a closed module. 
    \item DiCE: This approach can handle categorical features by user selection. More interpretable results are obtained by using the scaling, one hot encoding and model as a closed block. 
\end{itemize}

\subsection{Examples}

To demonstrate the application of the presented explainability techniques, we present a few examples using the Home Credit dataset (Sec.~\ref{sec:datasets}) and models previously trained (Sec.~\ref{sec:credit_models}). \giovani{To underscore the significance of explainability as a tool for model auditing, we examine models incorporating the sensitive attribute (aware models). Our aim is to determine whether the algorithm uses this attribute in assessing default risk.}

Our first example employs global explanations to compare Logistic Regression and Gradient Boosting. Two approaches are also considered to obtain global explanations: internal access to the model and global SHAP explanations. We enrich this example by comparing the explanations calculated with models fitted in all 10 folds. While LR and GBM feature importance can be obtained directly, calculating global explanations with SHAP is costly, as it is necessary to average local explanations obtained from multiple samples. In addition, to compare the SHAP values of multiple models, their output must be at the same scale, which is not necessarily true for models that have not been calibrated. For that reason, SHAP was utilized to explain the discrete predictions and not the probabilities. Fig.~\ref{fig:global_importances} presents parallel coordinate plots~\cite{inselberg2009parallel} with the features importance for the ten most important ones. In each plot, a line represents a fold, and the importance is represented at the horizontal position of each axis. The displayed features are those with the highest median importance among the 10 folds. 

It is possible to see that the importance of the features was similar among the 10 folds, in particular for the Logistic model, with almost the same coefficient values in all fitted models. In contrast, the number of splits of Gradient-Boosting presents a high variance. This occurs because it is a complex model that is more sensitive to the training data and each fold will result in trees that are very different. This is one advantage of simple models such as the Logistic one. We can also identify that many features are present in all of the plots. For example, \texttt{EXT\_SOURCE\_1}, \texttt{EXT\_SOURCE\_2}, \texttt{EXT\_SOURCE\_3} are individuals' score values imported from other credit institutions and are important for all the models. The external scores summarize many important attributes of the individual in terms of their payment capabilities. Their importance for the default risk was also identified by \citet{bucker2022transparency} and is in accordance with the knowledge of the domain.

Considering the Logistic Regression coefficients, features \texttt{CODE\_GENDER=M} and \texttt{CODE\_GENDER=F} were among the most important. In particular, the coefficient for \texttt{CODE\_GENDER=M} is positive, indicating that if the value of this binary feature is equal to $1$, the probability of default increases. The feature coefficient for \texttt{CODE\_GENDER=F} is negative. \texttt{CODE\_GENDER=M} was also an important feature for Gradient-boosting. \giovani{This can motivate concerns about the fairness of the trained models. This was investigated at Sec.~\ref{sec:fairness}, which highlighted that the \textit{aware} models presented significant bias towards an specific gender.}

\figGlobalExp

Global explanations obtained from the coefficients of the Logistic model are the only ones presented that can have a negative sign. This sign is important for analysis, as a feature with negative coefficients indicates that increasing this feature will decrease the probability of a positive outcome. Despite local SHAP explanations also presenting interpretable signs, the aggregation of local explanations causes the loss of such information. 

Despite Gradient Boosting having information on the number of splits, describing how much each feature was utilized during the training step, yet, it is not a value that represents the relation of such feature and the prediction of the final model. In Fig.~\ref{fig:global_importances} it is possible to see that SHAP values for the Gradient-Boosting model have a smaller variance compared to the number of splits and share many features with the global explanations of the LR model. It is interesting to see that \texttt{DAYS\_BIRTH}, a measure of the individual's age, was the second most important feature considering the number of splits, yet it is the eighth most important feature obtained with SHAP values. 

In the following examples, we analyze only the first fold for simplicity. Fig.~\ref{fig:pdp_ice} presents the partial dependence plot and the individual conditional expectation plot of the feature \texttt{AMT\_CREDIT} (credit amount requested) with the four models trained from Sec.~\ref{sec:credit_models}. Higher values of requested credit are expected to result in a higher probability of default, as more effort is needed to fulfill the payment. This relation is learned by all models considered, with small differences. Due to the model formulation, the relation is linear for the Logistic model, and the Neural Network also learned a linear relation. The Random Forest classifier presents a different pattern: the average probability decreases when the credit amount is over $5\cdot 10^5$. The Gradient-Boosting model also presents this increase in the probability of default as the requested credit amount increases. However, this effect is reduced when the requested amount is already large. We can analyze this relation in more detail with the ICE plot of 100 random samples. Individual lines permit better comprehension of the relationship between the feature and the model. Looking at the ICE plot of the Random Forest classifier, only a few samples presented a decrease in the probability of default decreases as the amount of requested credit increases. One possibility for such a characteristic is an interaction between \texttt{AMT\_CREDIT} and another feature, such as salary. \giovani{To verify this hypothesis, lines were colored based in the income of each client (\texttt{AMT\_INCOME\_TOTAL}), darker colors indicate a higher income. The samples in which the default risk decreases as the requested credit increases are samples with high income, indicating that the Random Forest classifier learned such relation.}

\figPDPICE

We now use SHAP and LIME to obtain local explanations for predicting the gradient-boosting model. With local explanations, we can verify in greater detail the decision process of the black-box model, and considering the credit scoring application, we can explain to a client why the algorithm predicted the default. Fig.~\ref{fig:shap_lime} presents the SHAP and LIME local explanations for the same individual. The bar plot presents the 7 features with the highest absolute importance, and the table presents the values of such features. This example was selected due to the high importance of gender in the final prediction. Considering the SHAP local explanation, \texttt{CODE\_GENDER} (with a value equal \texttt{M}) has the highest attribution among features, followed by \texttt{AMT\_GOODS\_PRICE}, that is, the price of goods involved in the loan. This was not observed in the LIME explanations, where the most important features were similar to the ones obtained from the global explanations. LIME identified that in the neighborhood of the selected sample, increasing \texttt{EXT\_SOURCE\_3, EXT\_SOURCE\_2, AMT\_GOODS\_PRICE, EXT\_SOURCE\_1} decreases the probability of default, while the observed categories of \texttt{CODE\_GENDER} and \texttt{NAME\_CONTRACT\_TYPE} and increasing \texttt{AMT\_CREDIT} increase the probability of default. The disagreement of explanation techniques is still an important topic of research~\cite{solunke2024mountaineer}. This example again highlights the utility of explainability for analyzing the algorithm's fairness. A greater analysis is necessary to understand why such great importance is calculated for the gender feature and approaches to reduce it should be considered, as the ones presented in Sec.~\ref{sec:fairness}. 

\figSHAPLIME

Lastly, we exemplify the use of explainability techniques designed to generate counterfactual explanations, i.e., with a sample with an undesired outcome, how one can alter feature values to achieve the desired one. Notice that local explanations and counterfactual explanations can be used similarly, however, counterfactual explanations are designed considering that the provided information should help the individual to obtain the desired outcome in future evaluation of the model. Considering that, counterfactual explanations should not include features that can not be easily altered by the client, such as gender, number of children, and others. In this example, we utilize the Logistic model to generate counterfactual explanations. The desired outcome is obtaining the prediction $\hat Y = 0$, and 24 features were selected to create the counterfactuals, including features regarding the client's documentation at the bank, characteristics of its home, amount of credit requested, and others. The most critical features, \texttt{EXT\_SOURCE\_1, EXT\_SOURCE\_2, EXT\_SOURCE\_3}, were not included in the generation of counterfactuals. 

MAPOCAM was executed to optimize for two costs of a counterfactual explanation: the number of features changed and the closeness of the counterfactual to the individual. The search considered only altering at most five features simultaneously. Tab.~\ref{table:mapocam} presents the obtained counterfactuals. Tab.~\ref{table:dice} presented the counterfactuals generated by the Dice algorithm. It can generate an arbitrary number of counterfactuals, and for comparison, we selected to generate the same number of counterfactuals as those found by MAPOCAM. The remaining parameters were set as default. Due to the search-based approach of MAPOCAM, it has the guarantee of identifying the optimal solutions, i.e., counterfactuals that minimize both of the costs. This advantage is obtained with a higher computational cost. DiCE can generate counterfactuals faster. However, the solutions identified tend to be more distant to the individual. 

\begin{table}
\centering
\begin{tabular}{lllll}
\hline
 & Original & CF 1 & CF 2 & CF 3 \\ \hline
 \texttt{AMT\_CREDIT} & $277\cdot 10^3$ &	---	& $255\cdot 10^3$ &	$255 \cdot 10^3$ \\
 \texttt{FLAG\_EMAIL} & 0 &	1 &	1 &	1 \\
\texttt{REG\_REGION\_NOT\_LIVE\_REGION} & 0 &	1 &	1 &	1 \\
 \texttt{BASEMENTAREA\_AVG} & 0.088 &	0.293 &	0.279 &	0.249 \\
 \texttt{DAYS\_LAST\_PHONE\_CHANGE} & -1438&	--- &	--- &	-1727  \\ \hline
\end{tabular}
\caption{Counterfactuals generated for a individual with MAPOCAM algorithm.}
\label{table:mapocam}
\end{table}

\begin{table}
\centering
\begin{tabular}{lllll}
\hline
 & Original & CF 1 & CF 2 & CF 3 \\ \hline
 \texttt{AMT\_INCOME\_TOTAL} &$0.2 \cdot 10^6$	&---	 & $12 \cdot 10^6$ &	$13 \cdot 10^6$\\ 
 \texttt{BASEMENTAREA\_AVG} & 0.088 &	0.666	 &---&	---\\ 
 \texttt{LIVINGAREA\_AVG} & 0.107 &--- &	--- &	0.424 \\
 \hline
\end{tabular}
\caption{Counterfactuals generated for a individual with DiCE algorithm.}
\label{table:dice}
\end{table}

In this example, all solutions from MAPOCAM identified that the user should register their e-mail within the bank and live in the region of the registry within the bank region. Two counterfactual explanations asked for a small reduction in the amount of credit requested. All counterfactuals also asked for a larger basement, and one of the counterfactuals identified that the user changed the cell phone too close to the application. Notice that the three counterfactuals identified by DiCE utilize less features, however they ask for really drastic alterations. An increase of ten times the client's income and an increase of almost eight in the basement area or four times an increase in the living area is asked for.

In credit scoring applications, explainability methods permit practitioners to gain a deeper understanding of the inner workings of machine learning models. By providing insights into the decision-making process, these techniques enable us to identify and rectify potential biases that may inadvertently lead to discriminatory outcomes. Moreover, explainability empowers clients by elucidating the impact of their features on loan approval, fostering transparency and understanding in an otherwise opaque process. The ability to debug models, identify discriminatory biases, and enhance client comprehension underscores the critical importance of explainability in building responsible and trustworthy credit scoring systems.

\section{Conclusion}

Banking institutions can leverage machine learning algorithms to achieve more accurate credit scoring systems. However, decisions pose a high impact on the individuals' conditions and these algorithms must be developed under responsible considerations. Without further attention, such algorithms can present hidden pitfalls, reinforce biases and discrimination against historically marginalized populations. In this work, we introduced the areas of fairness, reject inference, and explainability and their applications to the credit scoring problem. Under each topic, we present a set of representative techniques that cover diverse usage scenarios. A set of examples was also presented so that the reader can obtain theoretical and practical knowledge. The research in trustworthy machine learning algorithms is still a new topic and of great relevance. The practical applications of these approaches can lead to improvements in credit scoring and concession systems, making them more transparent and accessible. Future directions include the study of algorithms when applied over a long period and when the sensitive attribute is not known. Explainability techniques only increase their relevance as more complex models are trained, such as foundation models.

\section{Acknowledgements}

This project was supported by the brazilian Ministry of Science, Technology and Innovations, with resources from Law nº 8,248, of October 23, 1991, within the scope of PPI-SOFTEX, coordinated by Softex and published Arquitetura Cognitiva (Phase 3), DOU 01245.003479/2024 -10



\bibliographystyle{unsrtnat}






\end{document}